\pdfoutput=1
\documentclass[journal]{IEEEtran}
\usepackage{amsmath,amsfonts}
\usepackage{algorithmic}
\usepackage{algorithm}
\usepackage{array}
\usepackage[caption=false,font=normalsize,labelfont=sf,textfont=sf]{subfig}
\usepackage{textcomp}
\usepackage{stfloats}
\usepackage{url}
\usepackage{verbatim}
\usepackage{graphicx}
\usepackage{cite}
\usepackage{booktabs}
\usepackage{multirow}
\usepackage{xcolor}
\usepackage{tikz}
\usetikzlibrary{svg.path}
\usepackage[hidelinks]{hyperref}
\hypersetup{pdftitle={Progressive Multi-Scale Super-Resolution via Cascaded Bicubic Anchors and Inter-Scale Mutual Refinement},pdfauthor={Chun-An Lin, Yu-Chia Chang, Chi-Mao Fan, Yen-Chieh Ouyang, and Tsung-Jung Liu}}
\definecolor{orcidlogocol}{HTML}{A6CE39}
\tikzset{orcidlogo/.pic={%
 \fill[orcidlogocol] svg{M256,128c0,70.7-57.3,128-128,128C57.3,256,0,198.7,0,128C0,57.3,57.3,0,128,0C198.7,0,256,57.3,256,128z};
 \fill[white] svg{M86.3,186.2H70.9V79.1h15.4v48.4V186.2z}
   svg{M108.9,79.1h41.6c39.6,0,57,28.3,57,53.6c0,27.5-21.5,53.6-56.8,53.6h-41.8V79.1z M124.3,172.4h24.5c34.9,0,42.9-26.5,42.9-39.7c0-21.5-13.7-39.7-43.7-39.7h-23.7V172.4z}
   svg{M88.7,56.8c0,5.5-4.5,10.1-10.1,10.1c-5.6,0-10.1-4.6-10.1-10.1c0-5.6,4.5-10.1,10.1-10.1C84.2,46.7,88.7,51.3,88.7,56.8z};}}
\newsavebox{\orcidbox}
\savebox{\orcidbox}{\begin{tikzpicture}[yscale=-1,transform shape]\pic{orcidlogo};\end{tikzpicture}}
\newcommand{\orcid}[1]{\href{https://orcid.org/#1}{\raisebox{-0.15ex}{\resizebox{!}{1.6ex}{\usebox{\orcidbox}}}}}
\usetikzlibrary{arrows.meta,positioning,calc,fit}

\begin{document}

\title{Progressive Multi-Scale Super-Resolution via Cascaded Bicubic Anchors and Inter-Scale Mutual Refinement}

\author{Chun-An~Lin\,\orcid{0009-0004-8874-6196},~\IEEEmembership{Student~Member,~IEEE,} Yu-Chia Chang, Chi-Mao Fan, Yen-Chieh~Ouyang\,\orcid{0000-0002-4221-2787},~\IEEEmembership{Life~Member,~IEEE,} and Tsung-Jung~Liu\,\orcid{0000-0003-4296-0942},~\IEEEmembership{Member,~IEEE}
\thanks{This work has been submitted to the IEEE for possible publication. Copyright may be transferred without notice, after which this version may no longer be accessible.}%
\thanks{This work was supported in part by the National Science and Technology Council, Taiwan, under Grant NSTC 114-2221-E-005-040-MY2. \textit{(Corresponding author: Tsung-Jung Liu.)}}%
\thanks{C.-A. Lin is with the Department of Electrical Engineering, National Chung Hsing University, Taichung 402202, Taiwan, and also with the College of Computing and Informatics, Providence University, Taichung 433303, Taiwan.}%
\thanks{Y.-C. Chang, Y.-C. Ouyang, and T.-J. Liu are with the Department of Electrical Engineering, National Chung Hsing University, Taichung 402202, Taiwan (e-mail: tjliu@email.nchu.edu.tw).}%
\thanks{C.-M. Fan is with ASUSTeK Computer Inc., Taipei 112019, Taiwan.}}

\markboth{Preprint submitted to the IEEE}%
{Lin \MakeLowercase{\textit{et al.}}: Progressive Multi-Scale Super-Resolution via Cascaded Bicubic Anchors and Inter-Scale Mutual Refinement}

\setlength{\textfloatsep}{8pt plus 2pt minus 3pt}
\setlength{\dbltextfloatsep}{8pt plus 2pt minus 3pt}
\setlength{\floatsep}{9pt plus 2pt minus 2pt}
\setlength{\dblfloatsep}{9pt plus 2pt minus 2pt}
\AtBeginDocument{%
  \setlength{\abovedisplayskip}{5pt plus 2pt minus 2pt}%
  \setlength{\belowdisplayskip}{5pt plus 2pt minus 2pt}%
  \setlength{\abovedisplayshortskip}{3pt plus 1pt}%
  \setlength{\belowdisplayshortskip}{3pt plus 1pt}%
}

\maketitle
\bstctlcite{IEEEexample:BSTcontrol}

\begin{abstract}
Super-resolution networks are trained for one magnification factor, so several output resolutions require several models. We propose PMS-SR, a progressive single-input multiple-output network that produces $\times2$, $\times4$, and $\times8$ results in one forward pass, with a Transformer backbone on the low-resolution grid. In Stage~1, each scale is predicted as a residual over the detached bicubic upsampling of the previous scale's prediction. With Dirac and zero initialization, training starts exactly from the bicubic cascade, and each head adds only the band the previous scale lacks. In Stage~2, inter-scale mutual refinement couples the scales in both directions through space-to-depth feedback, deformable alignment, and per-pixel fusion, so $\times8$ features refine the $\times2$ output. Trained on DF2K, the 7.95M-parameter model achieves a higher PSNR than every compared multi-scale method at $\times4$ and $\times8$ on all five standard benchmarks. At $\times8$, it outperforms the $\times8$-specific DRN-L on all of them, leads all compared methods on Urban100, and is within 0.13\,dB of the best result elsewhere. Against three single-scale SwinIR-M models, PMS-SR uses 22\% of the parameters, runs $2.9\times$ faster on BSD100 inputs, and its outputs agree 6.7--12.4\,dB more closely. Code and pretrained models will be released at \url{https://github.com/julian135707/PMS-SR} upon acceptance.
\end{abstract}


\section{Introduction}
\label{sec:intro}
\IEEEPARstart{S}ingle-image super-resolution (SISR) reconstructs a high-resolution (HR) image from a low-resolution (LR) input. It is used throughout the video pipeline, in surveillance, adaptive streaming, zoomable playback, and content remastering, as well as in medical imaging \cite{wang2020deep} and thermal infrared imaging \cite{lin2024lirsrn,lin2026tirmamba}. Since SRCNN \cite{dong2016image}, accuracy has improved through deeper residual networks such as EDSR \cite{lim2017enhanced}, attention-based networks such as RCAN \cite{zhang2018image}, and, more recently, window-attention Transformers such as SwinIR \cite{liang2021swinir} and HAT \cite{chen2023activating}.

Most of these networks are single-scale, with one model per magnification factor. Many video systems, however, need several renditions of the same content, for example the resolution ladders of adaptive streaming \cite{menon2024jtps,amirpour2025deepstream}, zoomable playback, and surveillance viewers that switch between an overview and a detailed view. With single-scale models, each rendition needs its own network or is downsampled from the largest one. Multi-scale SISR methods share this computation. LapSRN and MS-LapSRN \cite{lai2017deep,lai2019fast} predict a Laplacian pyramid progressively, MSWSR \cite{zhang2020multi} and MLWAN \cite{ma2022mlwan} produce multi-scale outputs in the wavelet domain, and ProSR \cite{wang2018prosr} grows an asymmetric dense pyramid with strong $\times8$ results. These works show that progressive $\times2$ steps extend well to $\times8$ and that the way the scales are coupled, through shared features and the residual anchor of each scale, largely determines multi-scale accuracy. These multi-output methods, however, are convolutional, spend much of their computation on upsampled grids, and pass information only from coarse to fine, so fine-scale evidence never corrects their coarser outputs. More recent studies refine progressive and closed-loop structures \cite{wu2023fpan,huang2023clsr} or strengthen the single-scale backbone \cite{huang2024como,xie2025mat}, but they still train one model per factor, and arbitrary-scale methods \cite{chen2021learning,zhang2026cgssr} run one decoding pass per requested scale.

In this paper, we treat $\times2/{\times}4/{\times}8$ super-resolution (SR) as one coupled reconstruction problem. All deep feature extraction runs on the LR grid with cross-shaped window (CSWin) attention \cite{dong2022cswin}, whose horizontal and vertical stripes cover both image axes at low cost, complemented by the channel attention block of HAT \cite{chen2023activating}. In Stage~1, a pyramid with decreasing channel width expands the features $\times2\!\to\!\times4\!\to\!\times8$ by sub-pixel convolution \cite{shi2016real} and, in the spirit of MS-LapSRN \cite{lai2019fast}, predicts each scale as a residual over the bicubic upsampling of the previous scale's prediction, so each head only adds the band that the previous scale cannot represent. In Stage~2, inter-scale mutual refinement (IMR) refines all scales jointly: fine features are folded back to coarser grids by space-to-depth, coarse features are deformably aligned to finer grids, and a per-pixel fusion weights the streams; the $\times8$ refinement runs on the $\times4$ grid at one quarter of the cost.

Stripe attention, channel attention branches, sub-pixel upsampling, deformable warping, and residual multi-stage refinement are established components, and we reuse them unchanged. Our contribution lies in the mechanisms that couple the scales in the single-input multiple-output (SIMO) setting. The main contributions are as follows.
\begin{itemize}
    \item \textbf{A SIMO formulation of $\times2/{\times}4/{\times}8$ SR:} a single 7.95M-parameter model, smaller than any single-scale model in our comparison (Sec.~\ref{sec:quant}), produces all three outputs in one pass, and all of its attention runs on the LR grid.
    \item \textbf{Cascaded residual anchoring:} each scale is anchored on the detached, bicubically upsampled prediction of the previous scale. With Dirac and zero initialization, training starts exactly from the bicubic cascade, and each higher head adds only the band that the previous scale lacks instead of the increasingly large residual left by a direct anchor. Replacing the cascade with a direct anchor costs 2.4\,dB at $\times8$ on BSD100 in our ablation.
    \item \textbf{Inter-scale mutual refinement (IMR):} a second stage that combines space-to-depth feedback, deformable cross-scale alignment, and per-pixel fusion with a refine-then-upsample $\times8$ path. Fine-to-coarse signals have been used in SR before, but only inside single-scale models (Sec.~\ref{sec:relatedwork}); here they flow between renditions that are delivered together, so $\times8$ features also refine the $\times2$ output.
    \item \textbf{Accuracy and efficiency:} trained on DF2K and evaluated under the standard Y-channel protocol, PMS-SR has the highest PSNR among the compared multi-scale methods in every $\times4$ and $\times8$ entry. At $\times8$ it outperforms the $\times8$-specific DRN-L on all five standard benchmarks; against the strongest published $\times8$ models (SwinIR-M, DBPN-RES-MR64-3, and RDRN, with 1.5 to 4.6 times as many parameters), it is at least 0.16\,dB ahead on Urban100, significantly so against SwinIR-M in a paired test (adjusted $p<10^{-8}$), and at most 0.13\,dB behind elsewhere. Against three SwinIR-M models, it uses 22\% of the parameters and runs $2.9\times$ faster on BSD100 inputs.
    \item \textbf{Cross-scale consistency with degenerate controls:} we measure whether the delivered renditions agree, a property that per-factor evaluation does not capture. They agree 6.7--12.4\,dB more closely than those of three independently trained models, and because agreement alone is maximized by degenerate ladders, we report it together with the fidelity of each rendition.
\end{itemize}

\section{Related Work}
\label{sec:relatedwork}

\subsection{CNN-Based Super-Resolution}
SRCNN \cite{dong2016image} pioneered end-to-end SISR. EDSR \cite{lim2017enhanced} scaled up residual networks, RCAN \cite{zhang2018image} added channel attention, and D-DBPN \cite{haris2018deep} iterated up- and down-projections; its extended version \cite{haris2021dbpn} remains one of the strongest $\times8$ models. Sub-pixel convolution \cite{shi2016real} moved most of the computation to the LR grid and is now the standard upsampler. More recent convolutional networks add further priors or adapt to the input: SFHN \cite{wu2023sfhn} adds a Fourier-domain branch to its convolution blocks, the network in \cite{zuo2023jtf} lets image and gradient features guide each other, KCI-Net \cite{liu2026kcinet} predicts the contourlet coefficients of the HR image, and CoMoNet \cite{huang2024como} modulates its convolution kernels for each input image. Differentiable architecture search has also been used to find lightweight SR networks \cite{huang2023dlsr}. SFHN and CoMoNet-L train one model per factor, each larger than the whole PMS-SR (Sec.~\ref{sec:quant}).

\subsection{Transformer and State-Space Models}
SwinIR \cite{liang2021swinir} brought window attention to image restoration. Later Transformers such as EDT \cite{li2021efficient}, SRFormer \cite{zhou2023srformer}, HAT \cite{chen2023activating}, and MAT \cite{xie2025mat}, and state-space models such as MambaIR \cite{guo2024mambair}, MambaIRv2 \cite{guo2025mambairv2}, and SRMamba-T \cite{liu2025srmamba}, are also single-scale. Each of their per-factor models compared in Sec.~\ref{sec:quant} is larger than the whole PMS-SR. Other recent designs work in the frequency domain or reduce the cost: WaveletMixer \cite{zhang2026waveletmixer} combines the wavelet transform with self-attention, TRAN \cite{zheng2026tran} splits textures by complexity in the frequency domain and models them with separate Mamba experts, the lightweight PCTN \cite{li2025pctn} and FSPPTN \cite{yang2026fspptn} rely on pyramid clustering attention and frequency selection, respectively, and MattenSR \cite{dong2026mattensr} co-designs a Transformer--Mamba hybrid with its hardware accelerator. These designs also serve one factor per model, whereas PMS-SR runs its attention once on the LR grid for all three factors. CSWin \cite{dong2022cswin} pairs full-axis horizontal and vertical stripes, and HAT showed that a parallel channel attention branch complements window attention. We combine CSWin stripes with the channel attention of HAT and use NAFNet blocks \cite{chen2022simple} in the shallow feature extraction and reconstruction modules.

\subsection{Progressive and Multi-Scale Super-Resolution}
MDSR \cite{lim2017enhanced} shares a feature extractor across $\times2$, $\times3$, and $\times4$ but produces only one scale per forward pass. LapSRN and MS-LapSRN \cite{lai2017deep,lai2019fast} predict Laplacian residuals progressively with deep supervision. MSSR \cite{jia2017single}, MSWSR \cite{zhang2020multi}, and MLWAN \cite{ma2022mlwan} output multiple scales with compact convolutional or wavelet-based designs, and ProSR-L \cite{wang2018prosr} grows an asymmetric dense pyramid and reports the strongest published multi-scale $\times8$ results. PMS-SR differs from these methods in three respects: its stripe-window Transformer backbone operates only on the LR grid, its anchors form a bicubic cascade over predicted images rather than over the LR input, and a second stage couples the scales in both directions.

A related line serves many factors with one network through arbitrary-scale decoding. After implicit-function decoders such as LIIF \cite{chen2021learning}, CGSSR \cite{zhang2026cgssr} renders arbitrary-scale outputs with compact Gaussian splatting, OmniScaleSR \cite{chai2026omniscalesr} controls a diffusion prior by the target scale, and the method in \cite{zheng2026realarbvsr} extends arbitrary-scale SR to real-world video. These methods need one decoding pass per rendition. PMS-SR targets the opposite case: a fixed dyadic ladder whose three renditions are returned together and refined against each other. At $\times2$ and $\times4$, PMS-SR exceeds the 1.45M CGSSR$_{\mathrm{LTE}}$ in PSNR (Table~\ref{tab:multi_scale_comparison}).

Another line of work passes information from fine back to coarse, but within a single factor: D-DBPN \cite{haris2018deep} alternates up- and down-projections so that LR-grid errors correct the HR features, SRFBN \cite{li2019feedback} feeds a hidden state back to the input over unfolded iterations, and DRN \cite{guo2020closed} adds a dual regression that maps the SR output back to the LR input. FPAN \cite{wu2023fpan} links its stages by feedback connections and adds a pyramid non-local block; its 14.2M $\times4$ model is 0.16--0.70\,dB below PMS-SR (Table~\ref{tab:single_scale_comparison}). CLSR \cite{huang2023clsr} is the closest of these to our setting: for $\times4$ and $\times8$ it upsamples in $\times2$ steps and supervises the intermediate images, and a regression branch maps the gradient maps of the SR images back to the next lower scale, which closes the loop during training. In its forward path, however, information flows only from coarse to fine, the intermediate images serve as supervision, and a separate model of 14.4M--15.9M parameters is trained for each target factor (about 46M for $\times2$, $\times4$, and $\times8$). One 7.95M PMS-SR model serves all three and is comparable to CLSR at $\times4$ and $\times8$ (Sec.~\ref{sec:quant}). These methods show that fine-to-coarse signals help, but each targets one factor; IMR instead exchanges information between the three delivered renditions.

\subsection{Feature Alignment in Restoration}
EDVR \cite{wang2019edvr} popularized deformable alignment for video restoration, and DCNv2 \cite{zhu2019deformable} underlies most modern alignment modules. Later video SR methods align frames with attention: TCNet \cite{liu2023tcnet} learns inter-frame alignment with self-attention-based warping, CTVSR \cite{tang2024ctvsr} gathers information from adjacent frames by deformable cross attention, and the Transformer in \cite{zhang2023msvsr} corrects and aligns multi-scale features by cross attention. These methods align features across frames or feature levels; IMR instead aligns features from different output scales of the same image. The multi-stage MPRNet \cite{zamir2021multi} motivates the residual refinement in IMR; related designs include TSAN \cite{zhang2022tsan}, which reconstructs an SR image in two stages from coarse to fine, and the U-shaped attention module of \cite{lee2026ustm}, which refines the reconstruction stage of video SR networks. To this refinement, we add zero-initialized deformable alignment for cross-scale misalignment and a per-pixel softmax gate. In our test-time ablation, IMR improves all three outputs on every tested benchmark (Table~\ref{tab:ablation_imr}).

\begin{figure*}[!t]
\centering
\resizebox{0.76\textwidth}{!}{\input{figs/fig_arch_overview}}
\caption{Overall architecture of PMS-SR. All attention runs on the LR grid (blue). Stage~1 (orange) expands the features $\times2\!\to\!\times4\!\to\!\times8$ with decreasing widths; the $\times2$ output is a residual over the bicubically upsampled input, and each higher scale is a residual over the detached, bicubically upsampled prediction of the previous scale (dashed cascade). Stage~2, the IMR module (red, Fig.~\ref{fig:stage2}), refines all scales jointly; the heads of both stages are green. The bottom row shows the actual input and outputs of one forward pass at their true relative sizes.}
\label{fig:framework}
\end{figure*}

\section{Method}
\label{sec:method}

Given one LR input $I_{\mathrm{LR}}\in\mathbb{R}^{3\times H\times W}$, PMS-SR outputs six images in a single forward pass: the Stage-1 predictions $\hat{I}^{(1)}_{s}$ and the refined Stage-2 predictions $\hat{I}^{(2)}_{s}$ for $s\in\{2,4,8\}$. At test time, the Stage-2 outputs are used. As shown in Fig.~\ref{fig:framework}, the model comprises (i)~an LR-grid CSWin backbone, (ii)~a Stage-1 progressive reconstruction pyramid with cascaded anchors, and (iii)~the Stage-2 inter-scale mutual refinement (IMR) module. The input is reflection-padded to a multiple of~8 (the least common multiple of the stripe widths) and all outputs are cropped back, so arbitrary input sizes are supported.

\subsection{LR-Grid CSWin Backbone}
\label{sec:backbone}
A shallow feature extraction (SFE) module lifts the input to $C{=}128$ channels,
\begin{equation}
F_0=\mathrm{NAF}\bigl(\mathrm{Conv}_{3\times3}(I_{\mathrm{LR}})\bigr),
\end{equation}
where $\mathrm{NAF}(\cdot)$ is a NAFNet block \cite{chen2022simple} providing early channel mixing and simplified channel attention. Four residual CSWin groups follow. Each group contains four CSWin blocks and ends with a full $3{\times}3$ convolution plus a group-level residual, as in SwinIR \cite{liang2021swinir}. With $F_G$ denoting the output of the last group, the backbone output is
\begin{equation}
F_D=\mathrm{Conv}_{3\times3}\bigl(\mathrm{LN}(F_G)\bigr)+F_0,
\end{equation}
where $\mathrm{LN}$ is layer normalization. The global residual lets the attention layers focus on high-frequency content.

\begin{figure}[!t]
\centering
\resizebox{0.82\columnwidth}{!}{\begin{tikzpicture}[
    font=\small,
    box/.style={draw, rounded corners=1.5pt, minimum height=6.2mm, minimum width=21mm,
                align=center, inner sep=2pt},
    adopt/.style={box, fill=blue!10, draw=blue!45},
    ours/.style={box, fill=red!12, draw=red!60, very thick},
    plain/.style={box, fill=gray!8, draw=gray!55},
    sum/.style={circle, draw, inner sep=0.8pt},
    lbl/.style={font=\footnotesize},
    note/.style={font=\footnotesize, text=black!55, align=center},
    arr/.style={-{Stealth[length=1.7mm]}, semithick}
]
\node[plain, minimum width=14mm] (in)   at (0, 0)     {$X_{l-1}$};
\node[plain]                     (ln1)  at (0, -1.0)  {LN};
\node[adopt, minimum width=15.5mm] (ah) at (-2.0, -2.5) {H-stripe MSA\\{\footnotesize($C/2$, LePE)}};
\node[adopt, minimum width=15.5mm] (av) at (0.2, -2.5) {V-stripe MSA\\{\footnotesize($C/2$, LePE)}};
\node[adopt,  minimum width=13mm]   (cab) at (2.4, -2.5) {CAB};
\node[plain, minimum width=21mm] (proj) at (-0.9, -3.85) {Concat $+$ Proj};
\node[sum]                       (add1) at (0, -4.75) {$+$};
\node[plain]                     (ln2)  at (0, -5.7)  {LN};
\node[adopt]                     (leff) at (0, -6.7)  {LeFF};
\node[sum]                       (add2) at (0, -7.6)  {$+$};
\node[plain, minimum width=14mm] (out)  at (0, -8.45) {$X_{l}$};

\draw[arr] (in)  -- (ln1);
\draw[arr] (ln1.south) -- ++(0,-0.28) -| (ah.north);
\draw[arr] (ln1.south) -- ++(0,-0.28) -| (av.north);
\draw[arr] (ln1.south) -- ++(0,-0.28) -| (cab.north);
\draw[arr] (ah.south) -- ++(0,-0.35) -| ([xshift=-4mm]proj.north);
\draw[arr] (av.south) -- ++(0,-0.35) -| ([xshift=4mm]proj.north);
\draw[arr] (proj.south) -- ++(0,-0.3) -| (add1.north);
\draw[arr] (cab.south) -- node[right, lbl, pos=0.55] {$\times\,\alpha$} ++(0,-1.7) |- (add1.east);
\draw[arr] (in.west)  -- ++(-3.6,0) |- (add1.west);
\draw[arr] ($(add1.south)!0.35!(ln2.north)$) -- ++(-2.0,0) |- (add2.west);
\draw[arr] (add1) -- (ln2);
\node[lbl, anchor=west, inner sep=1pt] at (0.12, -5.12) {$X'$};
\draw[arr] (ln2)  -- (leff);
\draw[arr] (leff) -- (add2);
\draw[arr] (add2) -- (out);

\node[note, anchor=south west] at ([xshift=1mm,yshift=0.5mm]cab.north) {HAT CAB\\learnable $\alpha$};
\node[note, anchor=south] at ([xshift=-11.5mm,yshift=0.5mm]ah.north) {single-precision\\logits (both)};

\begin{scope}[shift={(-3.3,-9.3)}]
  \node[adopt, minimum width=4mm, minimum height=3mm, inner sep=0pt] (k1) at (0,0) {};
  \node[right=1.2mm of k1, lbl, align=left, text width=71mm, anchor=west]
    {adopted from prior work: stripe MSA with LePE~\cite{dong2022cswin},
     CAB~\cite{chen2023activating}, LeFF~\cite{wang2022uformer}};
\end{scope}
\node[lbl, align=left, text width=84mm, anchor=north west] at (-3.5, -9.75)
  {LN: layer normalization; MSA: multi-head self-attention (\mbox{H-/V-stripe} MSA: $\mathrm{Attn}_h$ and $\mathrm{Attn}_v$ in Eq.~\eqref{eq:block}); LePE: locally-enhanced positional encoding;
   CAB: channel attention block (Conv$3{\times}3$ $\to$ GELU $\to$ Conv$3{\times}3$ $\to$ SE);
   LeFF: locally-enhanced feed-forward network; Proj: linear projection;
   $\alpha$: learnable scalar (init.\ $0.01$). The contributions of this paper are in
   Stages~1 and~2 (Figs.~\ref{fig:framework} and~\ref{fig:stage2}).};
\end{tikzpicture}}
\caption{One CSWin block of PMS-SR (Eqs.~\eqref{eq:attn}--\eqref{eq:leff}): two parallel stripe attentions (horizontal and vertical, half the channels each) plus a channel attention block (CAB) scaled by a learnable $\alpha$, followed by a LeFF feed-forward network. Blue boxes are adopted from prior work without structural change: stripe attention with LePE \cite{dong2022cswin}, the CAB of HAT \cite{chen2023activating}, and LeFF \cite{wang2022uformer}.}
\label{fig:block}
\end{figure}

\subsubsection{Cross-Shaped Window Attention}
Each CSWin block (Fig.~\ref{fig:block}) splits the channels in half. One half attends within horizontal stripes of size $s_w\times W$, and the other within vertical stripes of size $H\times s_w$, where the stripe width $s_w$ is $4$ in the first two groups and $8$ in the last two. Together, the two branches form a cross-shaped receptive field that spans both axes of the feature map in a single layer \cite{dong2022cswin}. Attention uses locally-enhanced positional encoding (LePE), which adds a depthwise $3{\times}3$ convolution of the values to the attention output:
\begin{equation}
\mathrm{Attn}(Q,K,V)=\mathrm{softmax}\!\left(QK^{\top}/\sqrt{d}\right)V+\mathrm{DWConv}(V),
\label{eq:attn}
\end{equation}
where $Q$, $K$, and $V$ are the query, key, and value matrices of one stripe and $d$ is the head dimension.

\subsubsection{Parallel Channel Attention Branch}
Stripe attention, like shifted-window attention, has no access to global channel statistics \cite{chen2023activating}. We therefore adopt the channel attention block (CAB) of HAT \cite{chen2023activating} without structural change. It consists of a $3{\times}3$ convolution that compresses the channels by a factor of three, a GELU, a $3{\times}3$ convolution that restores them, and squeeze-and-excitation (SE). HAT runs the CAB in parallel with shifted-window attention; we run it in parallel with the stripe attention of every block. We make only two changes: the branch scale $\alpha$ is a learnable scalar initialized to $0.01$, the value that HAT keeps fixed, and the squeeze-and-excitation uses a reduction ratio of 16 instead of 30:
\begin{align}
X' &= X_{l-1}+\mathrm{Proj}\bigl[\mathrm{Attn}_{h}\,\|\,\mathrm{Attn}_{v}\bigr]\bigl(\mathrm{LN}(X_{l-1})\bigr)\nonumber\\
&\quad +\alpha\,\mathrm{CAB}\bigl(\mathrm{LN}(X_{l-1})\bigr),\label{eq:block}\\
X_{l} &= X'+\mathrm{LeFF}\bigl(\mathrm{LN}(X')\bigr),\label{eq:leff}
\end{align}
where $X_{l-1}$ and $X_{l}$ are the input and output of the $l$-th block, $\mathrm{Attn}_{h}$ and $\mathrm{Attn}_{v}$ are the horizontal- and vertical-stripe multi-head self-attentions (H- and V-stripe MSA in Fig.~\ref{fig:block}), $\|$ denotes channel concatenation, $\mathrm{Proj}$ is a linear projection, and LeFF \cite{wang2022uformer} is a feed-forward network with an interleaved $3{\times}3$ depthwise convolution for locality.

\subsection{Stage 1: Progressive Reconstruction with Cascaded Anchors}
\label{sec:stage1}
Computation is cheapest on coarse grids, so the channel width decreases as the resolution grows: $(C_2,C_4,C_8)=(128,80,64)$. Features are expanded by $\times2$ sub-pixel convolution $\mathrm{PS}_{\uparrow2}(\cdot)$ \cite{shi2016real} (a $3{\times}3$ convolution followed by PixelShuffle; each instance has its own weights and output width):
\begin{align}
F_2&=\mathrm{PS}_{\uparrow2}(F_D),\nonumber\\
F_4&=\Phi_4\bigl[\mathrm{PS}_{\uparrow2}(F_2)\,\|\,\mathrm{ctx}_4(F_D)\bigr],\nonumber\\
F_8&=\Phi_8\bigl[\mathrm{PS}_{\uparrow2}(F_4)\,\|\,\mathrm{ctx}_8(F_D)\bigr],
\end{align}
where $\mathrm{ctx}_s(\cdot)$ is a $1{\times}1$ projection followed by bilinear resizing to the $\times s$ grid and $\Phi_s$ is a $1{\times}1$ fusion. This LR context injection gives the $\times4$ and $\times8$ branches a direct path to the backbone that bypasses accumulated error.

Each scale is predicted by a lightweight head $\mathcal{H}_s$ (one NAF block and a two-layer convolutional tail) as a residual over an explicit image-domain anchor:
\begin{align}
\hat{I}^{(1)}_{2}&=\mathcal{H}_2(F_2)+S_2\bigl(\mathrm{bic}_{\uparrow2}(I_{\mathrm{LR}})\bigr),\label{eq:cascade1}\\
\hat{I}^{(1)}_{4}&=\mathcal{H}_4(F_4)+S_4\bigl(\mathrm{bic}_{\uparrow2}(\mathrm{sg}(\hat{I}^{(1)}_{2}))\bigr),\\
\hat{I}^{(1)}_{8}&=\mathcal{H}_8(F_8)+S_8\bigl(\mathrm{bic}_{\uparrow2}(\mathrm{sg}(\hat{I}^{(1)}_{4}))\bigr),
\label{eq:cascade}
\end{align}
where $\mathrm{bic}_{\uparrow r}$ denotes bicubic $\times r$ interpolation \cite{keys1981cubic} (the PyTorch implementation), $\mathrm{sg}(\cdot)$ denotes the stop-gradient (detach) operation, and $S_s$ is a $3{\times}3$ skip convolution. Because the $\times4$ output is anchored on the predicted $\times2$ image, $\mathcal{H}_4$ only has to add the band that $\hat{I}^{(1)}_{2}$ cannot represent. Two choices are intended to stabilize the cascade. First, the anchor is detached; otherwise the $\times8$ loss would back-propagate into $\mathcal{H}_2$ and $\mathcal{H}_4$ and change the effective weights of the deep supervision; ProSR reports instability for anchors built on predicted outputs \cite{wang2018prosr}. Second, the skip convolutions $S_s$ are Dirac-initialized and the last layer of every head is zero-initialized, so $\hat{I}^{(1)}_{s}$ equals the bicubic cascade exactly at initialization. A direct anchor would start within $0.03$\,dB of this point; what the cascade changes is the residual that each head then learns, which ablation axis~(c) isolates.

\begin{figure*}[!t]
\centering
\resizebox{0.75\textwidth}{!}{\begin{tikzpicture}[
    font=\small,
    box/.style={draw, rounded corners=1.5pt, minimum height=6.5mm, minimum width=17mm,
                align=center, inner sep=2.5pt},
    feed/.style={draw, rounded corners=1.5pt, minimum height=5.5mm, align=center,
                 inner sep=2pt, font=\footnotesize},
    down/.style={feed, fill=cyan!14, draw=cyan!55},
    up/.style={feed, fill=orange!22, draw=orange!60},
    da/.style={feed, fill=red!8, draw=red!55},
    ctxb/.style={feed, fill=teal!14, draw=teal!55},
    scf/.style={box, fill=purple!30, draw=purple!70},
    plain/.style={box, fill=gray!10, draw=gray!55},
    sum/.style={circle, draw, inner sep=1pt},
    src/.style={font=\footnotesize, text=black!60},
    lbl/.style={font=\footnotesize},
    lane/.style={font=\footnotesize\bfseries, text=black!65},
    arr/.style={-{Stealth[length=1.8mm]}, semithick},
    farr/.style={-{Stealth[length=1.6mm]}, thin},
    dot/.style={circle, fill, inner sep=0pt, minimum size=3.2pt}
]
\fill[gray!5, rounded corners=3pt] (-1.75, 1.55) rectangle (13.8, -1.5);
\fill[blue!4, rounded corners=3pt] (-1.75, -1.55) rectangle (13.8, -4.2);
\fill[gray!5, rounded corners=3pt] (-1.75, -4.25) rectangle (13.8, -6.95);

\node[lane] at (-1.15, 0) {$\times2$};
\node[plain] (f2) at (0, 0) {$F_2$};
\node[down] (pu42) at (4.5, 0.9)  {PU$_{\downarrow2}$, $80{\to}128$};
\node[down] (pu82) at (4.5, -0.9) {PU$_{\downarrow4}$, $64{\to}128$};
\node[scf]  (scf2) at (7.8, 0) {SCF$_{2}$};
\node[sum]  (a2)   at (9.5, 0) {$+$};
\node[plain] (f2e) at (11.0, 0) {$F^{e}_{2}$};

\draw[arr] (f2) -- (scf2);
\node[src, left=3.5mm of pu42] (s42) {$F_4$};
\node[src, left=3.5mm of pu82] (s82) {$F_8$};
\draw[farr] (s42) -- (pu42);
\draw[farr] (s82) -- (pu82);
\draw[farr] (pu42.east) -- ++(0.55,0) |- ([yshift=1.9mm]scf2.west);
\draw[farr] (pu82.east) -- ++(0.55,0) |- ([yshift=-1.9mm]scf2.west);
\draw[arr] (scf2) -- (a2);
\draw[arr] (f2.south) -- ++(0,-1.05) -| (a2.south);
\draw[arr] (a2) -- (f2e);

\node[lane] at (-1.15, -2.75) {$\times4$};
\node[plain] (f4) at (0, -2.75) {$F_4$};
\node[up]   (ps4)  at (3.6, -1.85) {PS$_{\uparrow2}$, $128{\to}80$};
\node[da]   (uda4) at (5.75, -1.85) {DA$_{4}$};
\draw[farr] (ps4) -- (uda4);
\node[down] (pu84) at (4.5, -3.65) {PU$_{\downarrow2}$, $64{\to}80$};
\node[scf]  (scf4) at (7.8, -2.75) {SCF$_{4}$};
\node[sum]  (a4)   at (9.5, -2.75) {$+$};
\node[plain] (f4e) at (11.0, -2.75) {$F^{e}_{4}$};

\draw[arr] (f4) -- (scf4);
\node[dot] at (5.75, -2.75) {};
\draw[farr] (5.75, -2.75) -- (uda4.south) node[pos=0.45, right, lbl, inner sep=1.5pt] {ref};
\node[src, left=3.5mm of ps4] (su4) {$F^{e}_{2}$};
\node[src, left=3.5mm of pu84] (s84) {$F_8$};
\draw[farr] (su4) -- (ps4);
\draw[farr] (s84) -- (pu84);
\draw[farr] (uda4.east) -- ++(0.55,0) |- ([yshift=1.9mm]scf4.west);
\draw[farr] (pu84.east) -- ++(0.55,0) |- ([yshift=-1.9mm]scf4.west);
\draw[arr] (scf4) -- (a4);
\draw[arr] (f4.south) -- ++(0,-1.05) -| (a4.south);
\draw[arr] (a4) -- (f4e);

\node[lane] at (-1.15, -5.5) {$\times8$};
\node[plain] (f8) at (0, -5.5) {$F_8$};
\node[down, minimum width=17mm] (sq8) at (2.5, -5.5) {PU$_{\downarrow2}$, $64{\to}80$};
\node[da]   (da8)  at (4.95, -4.6) {DA$_{8}$};
\node[ctxb] (ctx8) at (4.95, -6.4) {ctx$^{e}_{8}$: $1{\times}1$, $128{\to}80$};
\node[scf]  (scf8) at (7.8, -5.5) {SCF$_{8}$};
\node[up, minimum width=16mm] (ps8) at (10.0, -5.5) {PS$_{\uparrow2}$, $80{\to}64$};
\node[sum]  (a8)   at (11.5, -5.5) {$+$};
\node[plain] (f8e) at (12.8, -5.5) {$F^{e}_{8}$};

\draw[arr] (f8) -- (sq8);
\draw[arr] (sq8) -- (scf8);
\node[lbl, anchor=south, inner sep=1.5pt] at (4.2, -5.5) {$\tilde{F}_8$};
\node[dot] at (4.95, -5.5) {};
\draw[farr] (4.95, -5.5) -- (da8.south) node[pos=0.45, right, lbl, inner sep=1.5pt] {ref};
\node[src, left=3.5mm of da8]  (sd8) {$F^{e}_{4}$};
\node[src, left=3.5mm of ctx8] (sc8) {$F_D$};
\draw[farr] (sd8) -- (da8);
\draw[farr] (sc8) -- (ctx8);
\draw[farr] (da8.east)  -- ++(0.55,0) |- ([yshift=1.9mm]scf8.west);
\draw[farr] (ctx8.east) -- ++(0.3,0) |- ([yshift=-1.9mm]scf8.west);
\draw[arr] (scf8) -- (ps8);
\draw[arr] (ps8) -- (a8);
\draw[arr] (f8.south) -- ++(0,-1.05) -| (a8.south);
\draw[arr] (a8) -- (f8e);
\node[lbl, text=black!60] at (6.0, -7.1)
    {entire $\times8$ refinement computed on the $\times4$ grid ($1/4$ cost)};

\begin{scope}[shift={(-0.6, -8.0)}, lbl/.append style={text height=1.5ex, text depth=0.25ex}]
  \node[down, minimum width=4mm, minimum height=3.2mm, inner sep=1pt] (k1) at (0,0) {};
  \node[right=1mm of k1, lbl] (t1) {space-to-depth feedback};
  \node[up, minimum width=4mm, minimum height=3.2mm, inner sep=1pt, right=4mm of t1] (k2) {};
  \node[right=1mm of k2, lbl] (t2) {upsampling};
  \node[da, minimum width=4mm, minimum height=3.2mm, inner sep=1pt, right=4mm of t2] (k3) {};
  \node[right=1mm of k3, lbl] (t3) {cross-scale alignment};
  \node[scf, minimum width=4mm, minimum height=3.2mm, inner sep=1pt, right=4mm of t3] (k4) {};
  \node[right=1mm of k4, lbl] (t4) {fusion};
  \node[ctxb, minimum width=4mm, minimum height=3.2mm, inner sep=1pt, right=4mm of t4] (k5) {};
  \node[right=1mm of k5, lbl] {context projection};
\end{scope}
\node[lbl, align=left, text width=140mm, anchor=north] at (6.4, -8.35)
  {PU$_{\downarrow r}$: PixelUnshuffle (space-to-depth) by factor $r$ followed by a $1{\times}1$ projection;
   PS$_{\uparrow2}$: $\times2$ sub-pixel upsampling; DA$_s$: deformable cross-scale alignment
   (zero-initialized offset and modulation predictors), ref marking its reference input; SCF$_s$: spatial--channel fusion
   (per-pixel softmax gate over three streams $+$ NAF block); ctx$^{e}_{8}$: context projected from $F_D$ and bilinearly resized to the $\times4$ grid.
   Gray unboxed labels give the source tensor of each feeder branch; all refinements are residual.};
\end{tikzpicture}}
\caption{The inter-scale mutual refinement (IMR) module, Stage~2 (Eqs.~\eqref{eq:imr2}--\eqref{eq:imr8}). Fine features are folded back by space-to-depth ($\mathrm{PU}_{\downarrow r}$), coarse features are aligned to the finer grid by deformable alignment ($\mathrm{DA}_s$), and the streams are fused per pixel ($\mathrm{SCF}_s$). The $\times8$ row refines entirely on the $\times4$ grid and expands only the correction (refine-then-upsample). All refinements are residual.}
\label{fig:stage2}
\end{figure*}

\subsection{Stage 2: Inter-Scale Mutual Refinement}
\label{sec:imr}
Stage~1 is strictly coarse-to-fine: errors at $\times2$ propagate to finer scales, and $\times8$ evidence never corrects the coarser ones. The IMR module (Fig.~\ref{fig:stage2}) runs one round of bidirectional exchange over the Stage-1 features $(F_2,F_4,F_8)$, with the backbone feature $F_D$ as extra context for the $\times8$ row.

\subsubsection{Space-to-Depth Feedback}
Fine-to-coarse feedback uses PixelUnshuffle (space-to-depth) by a factor $r$ followed by a $1{\times}1$ projection, denoted $\mathrm{PU}_{\downarrow r}$. The rearrangement is lossless: every sub-pixel phase becomes a separate channel, so, unlike pooling, no fixed low-pass filter discards high-frequency evidence before the learned projection:
\begin{equation}
F^{e}_{2}=F_2+\mathrm{SCF}_2\bigl[F_2,\;\mathrm{PU}_{\downarrow2}(F_4),\;\mathrm{PU}_{\downarrow4}(F_8)\bigr],
\label{eq:imr2}
\end{equation}
where the superscript $e$ marks the refined (Stage-2) features and $\mathrm{SCF}_s$ is the spatial--channel fusion described below.

\subsubsection{Deformable Cross-Scale Alignment}
Coarse-to-fine paths suffer sub-pixel misalignment after upsampling. A lightweight deformable alignment $\mathrm{DA}(\cdot,\cdot)$, similar to DCNv2 \cite{zhu2019deformable}, takes the feature to be warped as its first argument and the reference feature as its second. From the concatenated pair, two $3{\times}3$ convolutions predict one 2-D offset ($\tanh$-bounded to $\pm10$ pixels) and one modulation scalar per pixel. The first feature is then warped by bilinear sampling and fused with the reference by pointwise and depthwise convolutions. The last prediction layer is zero-initialized, giving zero offsets and unit modulation, so alignment starts as the identity mapping. With $\mathrm{DA}_4$ and $\mathrm{DA}_8$ denoting the two instances,
\begin{equation}
F^{e}_{4}=F_4+\mathrm{SCF}_4\bigl[F_4,\;\mathrm{DA}_4\bigl(\mathrm{PS}_{\uparrow2}(F^{e}_{2}),F_4\bigr),\;\mathrm{PU}_{\downarrow2}(F_8)\bigr].
\label{eq:imr4}
\end{equation}

\subsubsection{Refine-then-Upsample for \texorpdfstring{$\times8$}{x8}}
Refining directly on the $\times8$ grid is memory- and compute-intensive. We instead fold $F_8$ to a $\times4$-grid representation $\tilde{F}_8=\mathrm{PU}_{\downarrow2}(F_8)$, align $F^{e}_4$ to it, inject a backbone context stream $\mathrm{ctx}^{e}_{8}(F_D)$ (a separate $1{\times}1$ projection of $F_D$, bilinearly resized to the $\times4$ grid) as a low-frequency anchor, fuse on the $\times4$ grid, and expand only the correction back to $\times8$:
\begin{equation}
F^{e}_{8}=F_8+\mathrm{PS}_{\uparrow2}\Bigl(\mathrm{SCF}_8\bigl[\tilde{F}_8,\;\mathrm{DA}_8\bigl(F^{e}_{4},\tilde{F}_8\bigr),\;\mathrm{ctx}^{e}_{8}(F_D)\bigr]\Bigr).
\label{eq:imr8}
\end{equation}
This runs the $\times8$ refinement at one quarter of the cost.

\subsubsection{Spatial--Channel Fusion}
Each fusion operator $\mathrm{SCF}_s$ takes its three input streams separately and combines them with a per-pixel softmax gate. A $1{\times}1$ compression followed by a $3{\times}3$ convolution produces three spatial weight maps that sum to one at every pixel, and the weighted sum of the streams is refined by a NAF block. The gate lets each location decide how much to rely on each scale. Every PU, PS, DA, and SCF operator is a separate module with its own weights.

Finally, zero-initialized enhancement heads $\mathcal{H}^{e}_{s}$ (two NAF blocks and the same two-layer convolutional tail as $\mathcal{H}_s$) produce residual corrections in the image domain, following the multi-stage design of MPRNet \cite{zamir2021multi}:
\begin{equation}
\hat{I}^{(2)}_{s}=\hat{I}^{(1)}_{s}+\mathcal{H}^{e}_{s}(F^{e}_{s}),\qquad s\in\{2,4,8\}.
\end{equation}

\subsection{Training Objective}
\label{sec:loss}
All six outputs are deeply supervised. For a prediction $\hat{I}$ and its target $I$, the loss combines a Charbonnier term \cite{charbonnier1994two}, an edge term \cite{zamir2021multi}, and a frequency-domain term \cite{cho2021rethinking}:
\begin{equation}
\ell(\hat{I},I)=\ell_{\mathrm{char}}(\hat{I},I)+0.05\,\ell_{\mathrm{edge}}(\hat{I},I)+0.01\,\ell_{\mathrm{fft}}(\hat{I},I),
\label{eq:pairloss}
\end{equation}
where $\ell_{\mathrm{char}}=\frac{1}{N}\sum_{i}\sqrt{(\hat{I}_i-I_i)^2+\varepsilon^2}$ over the $N$ pixel values ($\varepsilon^2{=}10^{-6}$), $\ell_{\mathrm{edge}}$ is the $L_1$ distance between Laplacian filter responses, and $\ell_{\mathrm{fft}}$ is the $L_1$ distance between the real and imaginary parts of the orthonormal 2-D FFT coefficients. Frequency-domain supervision \cite{cho2021rethinking,jiang2021focal} and spatial--frequency processing \cite{lin2024infrared,lin2025lightweight} likewise emphasize high-frequency content, which a purely spatial loss tends to underweight. With scale weights $(\lambda_2,\lambda_4,\lambda_8)=(0.5,0.8,1.2)$ emphasizing the hardest factor, the total objective is
\begin{equation}
\begin{aligned}
\mathcal{L}^{(j)}&=\sum_{s\in\{2,4,8\}}\lambda_s\,\ell\bigl(\hat{I}^{(j)}_{s},I^{\mathrm{HR}}_{s}\bigr),\quad j\in\{1,2\},\\
\mathcal{L}&=\bigl(0.4\,\mathcal{L}^{(1)}+0.6\,\mathcal{L}^{(2)}\bigr)/2.5,
\end{aligned}
\label{eq:total}
\end{equation}
where the superscript $j$ indexes the stage: $\mathcal{L}^{(1)}$ supervises the Stage-1 outputs $\hat{I}^{(1)}_{s}$ and $\mathcal{L}^{(2)}$ the Stage-2 outputs $\hat{I}^{(2)}_{s}$. $I^{\mathrm{HR}}_{s}$ is the ground truth at scale $s$, i.e., the HR patch for $s=8$ and its bicubic downsampling by $8/s$ for $s=2$ and $4$, since training feeds the $\times8$ LR patch to all three outputs (Sec.~\ref{sec:datasets}). The divisor $2.5=\sum_s\lambda_s$ normalizes the scale weights, and the stage weights $0.4$ and $0.6$ favor the delivered Stage-2 outputs.

\section{Experiments}
\label{sec:Experiments}

\subsection{Datasets, Degradations, and Evaluation Protocol}
\label{sec:datasets}

\paragraph{Training sets}
Following recent SISR practice, we train on DF2K (DIV2K \cite{Agustsson_2017} and Flickr2K \cite{lim2017enhanced}, 3450 HR images). Each HR image is cropped to a multiple of 8 and downsampled by MATLAB-equivalent bicubic interpolation (with antialiasing) to $1/2$, $1/4$, and $1/8$ of its size. In every iteration, a random $48{\times}48$ LR patch and its aligned $96{\times}96$, $192{\times}192$, and $384{\times}384$ counterparts are cropped on the fly and augmented with the same random flips and $90^{\circ}$ rotations. The training and test degradations are identical.

\paragraph{Test benchmarks}
We report results on the five standard benchmarks, Set5 \cite{bevilacqua2012} (5 images), Set14 \cite{zeyde2010} (14 images), BSD100 \cite{martin2001} (100 images), Urban100 \cite{Huang-CVPR-2015} (100 images), and Manga109 \cite{aizawa2020building} (109 images), at $\times2$, $\times4$, and $\times8$, with LR inputs generated by bicubic downsampling in the MATLAB \texttt{imresize} convention \cite{keys1981cubic} after modulo cropping. Because SR papers use one image from each of the 109 Manga109 volumes, we checked that our copy is the standard set: its bicubic PSNRs at $\times2$ and $\times4$ agree within 0.02\,dB with those in \cite{zhang2018image}, and the released SwinIR-M model reproduces its published Manga109 scores exactly.

\paragraph{Evaluation protocol}
Following common practice \cite{zhang2018image,lai2019fast,wang2018prosr}, peak signal-to-noise ratio (PSNR) and structural similarity (SSIM) \cite{wang2004image} are computed on the Y (luminance) channel of YCbCr after conversion from sRGB, on 8-bit images with $s$ border pixels removed at scale $s$. At $\times8$, we also report the learned perceptual image patch similarity (LPIPS) \cite{zhang2018unreasonable} (AlexNet backbone, lower is better). To test whether a difference is systematic, we apply the two-sided Wilcoxon signed-rank test \cite{wilcoxon1945individual} to the per-image Y-PSNR differences on each benchmark and adjust the $p$-values with the Holm procedure \cite{holm1979simple} over the benchmarks of each comparison.

\subsection{Implementation Details}
\label{sec:impl}

\paragraph{Framework and hardware}
All experiments are implemented in PyTorch~2.2 (CUDA~12.1) and run on a single NVIDIA RTX~4090 GPU (24\,GB). Training the reported model takes about 146 GPU-hours, and gradient checkpointing on the $\times4$ and $\times8$ branches keeps the peak memory at 15.4\,GB. All random seeds are fixed.

\paragraph{Architecture hyperparameters}
The backbone uses an embedding width of $C{=}128$ and four groups of four CSWin blocks, with stripe widths $(4,4,8,8)$ and $(4,4,8,8)$ attention heads across the groups, a LeFF expansion ratio of $2.0$, $qkv$ bias, and stochastic depth with a drop rate that increases linearly over the blocks to a maximum of $0.1$.

\paragraph{Optimization}
All scales are trained jointly with the objective in \eqref{eq:total} for 300k iterations at batch size 16, using Adam ($\beta_1{=}0.9$, $\beta_2{=}0.999$) and a learning rate of $2{\times}10^{-4}$ with a 2k-iteration warm-up, halved at 150k, 240k, 270k, and 285k iterations. We train in bfloat16 with single-precision attention logits, clip gradients at 2.0, and keep an exponential moving average (EMA) of the weights with decay $\min(0.999,(1+t)/(10+t))$ at iteration $t$; the EMA weights are used for evaluation. The checkpoint with the best average full-image Y-PSNR on Set5 over the three scales (iteration 275k) is used. Since Set5 is also a test benchmark, we checked that this choice brings no advantage: the final 300k iterate scores $38.17/32.70/27.49$\,dB against $38.16/32.72/27.49$\,dB, and the three-scale average validation PSNR varies by only $0.009$\,dB (standard deviation) over the last 50k iterations.

\paragraph{Inference}
Test images are processed in $48{\times}48$ LR tiles with an overlap of 16 pixels, which matches the attention geometry seen in training, and all tables use this tiled protocol unless noted otherwise. Whole-image inference, which several baselines use, changes the PSNR of PMS-SR by at most 0.04\,dB at $\times8$ on the five benchmarks and by $-0.02$ to $+0.12$\,dB at $\times2$ and $\times4$ on Set5, Set14, BSD100, and Urban100 ($192{\times}192$ tiles for Urban100).

\begin{table*}[t]
\centering
\caption{Comparison with multi-scale super-resolution methods (PSNR\,/\,SSIM, Y channel). Best in \textcolor{red}{red}, second best in \textcolor{blue}{blue}; PMS-SR+ (self-ensemble) and the DIV2K-only control are not ranked. $^{\dagger}$\,=\,measured by us from the authors' released weights (Sec.~\ref{sec:quant}). CGSSR$_{\mathrm{LTE}}$ is an arbitrary-scale model, listed with its published PSNR.}
\label{tab:multi_scale_comparison}
\footnotesize
\renewcommand{\arraystretch}{1.0}
\setlength{\tabcolsep}{0pt}
\begin{tabular*}{\textwidth}{@{\extracolsep{\fill}} l c c ccccc @{}}
\toprule
Method & Scale & Params & Set5 & Set14 & BSD100 & Urban100 & Manga109 \\
 & & (M) & PSNR / SSIM & PSNR / SSIM & PSNR / SSIM & PSNR / SSIM & PSNR / SSIM \\
\midrule
\multicolumn{8}{l}{\textit{Scale} $\times 2$} \\
\midrule
LapSRN \cite{lai2017deep}   & $\times 2$ & 1.31 & 37.52 / 0.959 & 33.08 / 0.913 & 31.80 / 0.895 & 30.41 / 0.910 & 37.27 / 0.974 \\
MSWSR \cite{zhang2020multi} & $\times 2$ & 1.23 & 37.49 / 0.9582$^{\dagger}$ & 33.23 / 0.9123 & 31.88 / 0.8929 & 31.14 / 0.9169 & 37.32 / 0.9733 \\
MLWAN \cite{ma2022mlwan}    & $\times 2$ & 1.23 & 37.68 / 0.9589 & 33.47 / \textcolor{red}{\textbf{0.9227}} & 31.96 / 0.8974 & 31.34 / 0.9309 & 37.39 / 0.9733 \\
MS-LapSRN \cite{lai2019fast} & $\times 2$ & 0.22 & 37.78 / 0.960 & 33.28 / 0.915 & 32.05 / 0.898 & 31.15 / 0.919 & 37.78 / 0.976 \\
CGSSR$_{\mathrm{LTE}}$ \cite{zhang2026cgssr} & $\times 2$ & 1.45 & 37.96 / -- & 33.61 / -- & 32.18 / -- & 32.16 / -- & 38.52 / -- \\
MDSR \cite{lim2017enhanced} & $\times 2$ & 8.00 & 38.11 / 0.9602 & 33.85 / 0.9198 & 32.29 / \textcolor{blue}{\textbf{0.9007}} & 32.84 / 0.9347 & \textcolor{blue}{\textbf{39.09}} / \textcolor{red}{\textbf{0.9779}}$^{\dagger}$ \\
ProSR-L \cite{wang2018prosr} & $\times 2$ & 15.50 & \textcolor{red}{\textbf{38.17}} / \textcolor{blue}{\textbf{0.9608}}$^{\dagger}$ & \textcolor{red}{\textbf{34.00}} / \textcolor{blue}{\textbf{0.921}} & \textcolor{red}{\textbf{32.34}} / \textcolor{red}{\textbf{0.902}} & \textcolor{blue}{\textbf{32.91}} / \textcolor{blue}{\textbf{0.935}} & \textcolor{red}{\textbf{39.10}} / \textcolor{blue}{\textbf{0.9777}}$^{\dagger}$ \\
PMS-SR (DIV2K-only, $\tfrac{1}{3}$ sched.) & $\times 2$ & 7.95 & 38.04 / 0.9607 & 33.74 / 0.9186 & 32.17 / 0.8985 & 32.42 / 0.9310 & 38.57 / 0.9769 \\
\textbf{PMS-SR (Ours)}       & $\times 2$ & 7.95 & \textcolor{blue}{\textbf{38.16}} / \textcolor{red}{\textbf{0.9611}} & \textcolor{blue}{\textbf{33.95}} / 0.9200 & \textcolor{blue}{\textbf{32.31}} / 0.9002 & \textcolor{red}{\textbf{33.03}} / \textcolor{red}{\textbf{0.9366}} & 38.92 / \textcolor{blue}{\textbf{0.9777}} \\
\textbf{PMS-SR+ (Ours)}      & $\times 2$ & 7.95 & 38.21 / 0.9613 & 34.06 / 0.9208 & 32.35 / 0.9005 & 33.15 / 0.9375 & 39.07 / 0.9780 \\
\midrule
\multicolumn{8}{l}{\textit{Scale} $\times 4$} \\
\midrule
LapSRN \cite{lai2017deep}   & $\times 4$ & 1.31 & 31.54 / 0.885 & 28.19 / 0.772 & 27.32 / 0.728 & 25.21 / 0.756 & 29.09 / 0.890 \\
MSWSR \cite{zhang2020multi} & $\times 4$ & 1.23 & 32.00 / 0.8913$^{\dagger}$ & 28.47 / 0.7776 & 27.48 / 0.7311 & 25.78 / 0.7744 & 30.01 / 0.8999 \\
MLWAN \cite{ma2022mlwan}    & $\times 4$ & 1.23 & 32.19 / 0.8957 & 28.55 / 0.7797 & 27.67 / 0.7429 & 25.89 / 0.7779 & 30.16 / 0.9110 \\
MS-LapSRN \cite{lai2019fast} & $\times 4$ & 0.22 & 31.74 / 0.889 & 28.26 / 0.774 & 27.43 / 0.731 & 25.51 / 0.768 & 29.54 / 0.897 \\
CGSSR$_{\mathrm{LTE}}$ \cite{zhang2026cgssr} & $\times 4$ & 1.45 & 32.28 / -- & 28.56 / -- & 27.63 / -- & 26.17 / -- & 30.50 / -- \\
MDSR \cite{lim2017enhanced} & $\times 4$ & 8.00 & 32.50 / 0.8973 & 28.72 / 0.7857 & 27.72 / 0.7418 & 26.67 / 0.8041 & 31.04 / 0.9145$^{\dagger}$ \\
ProSR-L \cite{wang2018prosr} & $\times 4$ & 15.50 & \textcolor{blue}{\textbf{32.53}} / \textcolor{blue}{\textbf{0.8991}}$^{\dagger}$ & \textcolor{blue}{\textbf{28.94}} / \textcolor{blue}{\textbf{0.790}} & \textcolor{blue}{\textbf{27.79}} / \textcolor{blue}{\textbf{0.743}} & \textcolor{blue}{\textbf{26.89}} / \textcolor{blue}{\textbf{0.809}} & \textcolor{blue}{\textbf{31.39}} / \textcolor{blue}{\textbf{0.9181}}$^{\dagger}$ \\
PMS-SR (DIV2K-only, $\tfrac{1}{3}$ sched.) & $\times 4$ & 7.95 & 32.52 / 0.8985 & 28.86 / 0.7877 & 27.74 / 0.7420 & 26.77 / 0.8056 & 31.10 / 0.9159 \\
\textbf{PMS-SR (Ours)}       & $\times 4$ & 7.95 & \textcolor{red}{\textbf{32.72}} / \textcolor{red}{\textbf{0.9012}} & \textcolor{red}{\textbf{29.00}} / \textcolor{red}{\textbf{0.7915}} & \textcolor{red}{\textbf{27.87}} / \textcolor{red}{\textbf{0.7464}} & \textcolor{red}{\textbf{27.25}} / \textcolor{red}{\textbf{0.8177}} & \textcolor{red}{\textbf{31.69}} / \textcolor{red}{\textbf{0.9222}} \\
\textbf{PMS-SR+ (Ours)}      & $\times 4$ & 7.95 & 32.79 / 0.9020 & 29.08 / 0.7927 & 27.90 / 0.7470 & 27.38 / 0.8204 & 31.90 / 0.9237 \\
\midrule
\multicolumn{8}{l}{\textit{Scale} $\times 8$} \\
\midrule
LapSRN \cite{lai2017deep}   & $\times 8$ & 1.31 & 26.14 / 0.738 & 24.44 / 0.623 & 24.54 / 0.586 & 21.81 / 0.581 & 23.39 / 0.735 \\
MSWSR \cite{zhang2020multi} & $\times 8$ & 1.23 & 26.76 / 0.7631$^{\dagger}$ & 24.82 / 0.6338 & 24.74 / 0.5914 & 22.30 / 0.6045 & 24.21 / 0.7609 \\
MLWAN \cite{ma2022mlwan}    & $\times 8$ & 1.23 & 26.89 / 0.7732 & 24.97 / \textcolor{red}{\textbf{0.6574}} & 24.81 / 0.5969 & 22.95 / 0.6371 & 24.51 / 0.7833 \\
MS-LapSRN \cite{lai2019fast} & $\times 8$ & 0.22 & 26.34 / 0.753 & 24.57 / 0.629 & 24.65 / 0.592 & 22.06 / 0.598 & 23.90 / 0.759 \\
ProSR-L \cite{wang2018prosr} & $\times 8$ & 15.50 & \textcolor{blue}{\textbf{27.35}} / \textcolor{blue}{\textbf{0.7885}}$^{\dagger}$ & \textcolor{blue}{\textbf{25.29}} / 0.652 & \textcolor{blue}{\textbf{24.99}} / \textcolor{blue}{\textbf{0.606}} & \textcolor{blue}{\textbf{23.04}} / \textcolor{blue}{\textbf{0.645}} & \textcolor{blue}{\textbf{25.29}} / \textcolor{blue}{\textbf{0.8045}}$^{\dagger}$ \\
PMS-SR (DIV2K-only, $\tfrac{1}{3}$ sched.) & $\times 8$ & 7.95 & 27.26 / 0.7862 & 25.19 / 0.6490 & 24.99 / 0.6050 & 23.06 / 0.6448 & 25.13 / 0.7997 \\
\textbf{PMS-SR (Ours)}       & $\times 8$ & 7.95 & \textcolor{red}{\textbf{27.49}} / \textcolor{red}{\textbf{0.7922}} & \textcolor{red}{\textbf{25.37}} / \textcolor{blue}{\textbf{0.6551}} & \textcolor{red}{\textbf{25.07}} / \textcolor{red}{\textbf{0.6094}} & \textcolor{red}{\textbf{23.36}} / \textcolor{red}{\textbf{0.6593}} & \textcolor{red}{\textbf{25.58}} / \textcolor{red}{\textbf{0.8129}} \\
\textbf{PMS-SR+ (Ours)}      & $\times 8$ & 7.95 & 27.54 / 0.7935 & 25.44 / 0.6568 & 25.11 / 0.6103 & 23.46 / 0.6623 & 25.75 / 0.8158 \\
\bottomrule
\end{tabular*}
\end{table*}

\subsection{Quantitative Comparison with Representative Methods}
\label{sec:quant}

\paragraph{Comparison with multi-scale methods}
Table~\ref{tab:multi_scale_comparison} compares PMS-SR with the other multi-scale methods. Like PMS-SR, they serve several factors, so this is our primary comparison. Published values come from the original papers (ProSR SSIM from its supplementary material; MSWSR from the compilation in \cite{ma2022mlwan}). The cells marked $^{\dagger}$ fill unpublished entries with our measurements of the released MDSR model, the single released MSWSR checkpoint (for $\times2$, $\times4$, and $\times8$), and the released ProSR-L checkpoints on Set5 and Manga109. Our pipeline reproduces the published MSWSR results within 0.02\,dB, the result images released with ProSR-L score within 0.06\,dB of its published values, and its checkpoints reproduce the Set5 scores of those images within 0.02\,dB. The released MDSR weights, a PyTorch port, reproduce the published results only within $-0.12$ to $+0.09$\,dB, so the MDSR$^{\dagger}$ entries carry this uncertainty. ProSR-S is omitted because its weights are no longer available and ProSR-L is more accurate in every reported entry. PMS-SR+ uses the geometric self-ensemble \cite{lim2017enhanced}; like the DIV2K-only control (same architecture, 800 DIV2K images, one third of the schedule), it is shown for reference and not ranked.

At $\times8$, which MDSR cannot produce, PMS-SR is the best multi-scale method in every PSNR entry and in every SSIM entry except Set14, where MLWAN reports a higher SSIM (0.6574 vs.\ 0.6551) with a 0.40\,dB lower PSNR. PMS-SR is 0.08--0.32\,dB better than ProSR-L, which has about twice as many parameters, and 0.42--1.68, 0.26--1.07, and 0.33--1.37\,dB better than MS-LapSRN, MLWAN, and MSWSR, respectively. It also leads every column at $\times4$ (0.06--0.36\,dB above ProSR-L). At $\times2$, it leads on Urban100 and is within 0.05\,dB of ProSR-L on Set5, Set14, and BSD100, with the highest SSIM on Set5; on Manga109, ProSR-L$^{\dagger}$ and MDSR$^{\dagger}$ are 0.18 and 0.17\,dB ahead. The DIV2K-only control separates the effect of the training data. With one third of the schedule, it already comes close to ProSR-L at $\times8$ with half of its parameters ($+0.02$\,dB on Urban100, equal on BSD100, and $-0.09$, $-0.10$, and $-0.16$\,dB on Set5, Set14, and Manga109). At $\times8$ it also stays within 0.2\,dB of the DIV2K-trained RCAN and the DF2K-trained DRN-L on every benchmark (Table~\ref{tab:single_scale_comparison}). The design therefore accounts for most of the lead over MSWSR and MLWAN and reaches nearly the accuracy of ProSR-L at half its size, whereas the margin over ProSR-L also reflects the larger training set and the full schedule. Table~\ref{tab:multi_scale_comparison} also lists the arbitrary-scale CGSSR$_{\mathrm{LTE}}$ \cite{zhang2026cgssr}, a 1.45M model trained on DIV2K. PMS-SR is 0.13--0.87\,dB higher at $\times2$ and 0.24--1.19\,dB higher at $\times4$. With the same training set, our DIV2K-only control is 0.05--0.26\,dB higher at $\times2$ on four benchmarks and 0.01\,dB lower on BSD100, and 0.11--0.60\,dB higher at $\times4$. CGSSR$_{\mathrm{LTE}}$ accepts any factor with less than one fifth of the parameters; PMS-SR covers only the three dyadic factors and returns them in one pass.

\begin{table*}[!t]
\centering
\caption{Reference comparison with single-scale (SISO) models, each trained per factor (Params is the per-factor size), against one 7.95M PMS-SR covering all three factors in one pass ($^{\ast}$). Best in \textcolor{red}{red}, second best in \textcolor{blue}{blue}; PMS-SR+ (self-ensemble) is not ranked. $^{\dagger}$\,=\,measured by us from the authors' released weights (Sec.~\ref{sec:quant}); $\approx$\,=\,size reported for the $\times4$ model.}
\resizebox{\textwidth}{!}{%
\begin{tabular}{llcccccc}
\toprule
Method & Scale & Params & Set5 & Set14 & BSD100 & Urban100 & Manga109 \\
 & & (M) & PSNR / SSIM & PSNR / SSIM & PSNR / SSIM & PSNR / SSIM & PSNR / SSIM \\
\midrule
EDSR\cite{lim2017enhanced} (2017) & $\times2$ & 40.7 & 38.11 / 0.9601 & 33.92 / 0.9195 & 32.32 / 0.9013 & 32.93 / 0.9351 & 39.10 / 0.9773 \\
RCAN\cite{zhang2018image} (2018) & $\times2$ & 15.4 & 38.27 / 0.9614 & 34.12 / 0.9216 & 32.41 / 0.9027 & 33.34 / 0.9384 & 39.44 / 0.9786 \\
SAN\cite{dai2019second} (2019) & $\times2$ & 15.7 & 38.31 / 0.9620 & 34.07 / 0.9213 & 32.42 / 0.9028 & 33.10 / 0.9370 & 39.32 / 0.9792 \\
SwinIR-M\cite{liang2021swinir} (2021) & $\times2$ & 11.8 & 38.42 / 0.9623 & 34.46 / 0.9250 & 32.53 / 0.9041 & 33.81 / 0.9427 & 39.92 / 0.9797 \\
EDT\cite{li2021efficient} (2023) & $\times2$ & 11.5 & 38.45 / 0.9624 & 34.57 / 0.9258 & 32.52 / 0.9041 & 33.80 / 0.9425 & 39.93 / 0.9800 \\
SRFormer\cite{zhou2023srformer} (2023) & $\times2$ & 10.4 & 38.51 / 0.9627 & 34.44 / 0.9253 & 32.57 / 0.9046 & 34.09 / 0.9449 & 40.07 / 0.9802 \\
HAT\cite{chen2023activating} (2023) & $\times2$ & 20.8 & 38.63 / 0.9630 & 34.86 / 0.9274 & \textcolor{blue}{\textbf{32.62}} / \textcolor{blue}{\textbf{0.9053}} & 34.45 / 0.9466 & 40.26 / 0.9809 \\
FPAN\cite{wu2023fpan} (2023) & $\times2$ & 14.0 & 38.19 / 0.9612 & 33.88 / 0.9210 & 32.30 / 0.9012 & 32.72 / 0.9339 & 39.03 / 0.9772 \\
CLSR\cite{huang2023clsr} (2023) & $\times2$ & 15.9 & 38.41 / 0.9623 & 34.45 / 0.9247 & 32.48 / 0.9036 & 33.43 / 0.9398 & 39.63 / 0.9794 \\
SFHN\cite{wu2023sfhn} (2023) & $\times2$ & 21 & 38.33 / 0.9623 & 34.28 / 0.9223 & 32.44 / 0.9038 & 33.49 / 0.9390 & 39.90 / 0.9800 \\
CoMoNet-L\cite{huang2024como} (2024) & $\times2$ & 15.6 & 38.37 / 0.9620 & 34.52 / 0.9254 & 32.50 / 0.9039 & 33.68 / 0.9414 & 39.97 / 0.9801 \\
MambaIR\cite{guo2024mambair} (2024) & $\times2$ & 20.4 & 38.57 / 0.9627 & 34.67 / 0.9261 & 32.58 / 0.9048 & 34.15 / 0.9446 & 40.28 / 0.9806 \\
HMA\cite{chu2024hmanet} (2024) & $\times2$ & $\approx$69.9 & \textcolor{red}{\textbf{38.79}} / \textcolor{red}{\textbf{0.9641}} & \textcolor{red}{\textbf{35.11}} / \textcolor{red}{\textbf{0.9286}} & \textcolor{red}{\textbf{32.67}} / \textcolor{red}{\textbf{0.9061}} & \textcolor{red}{\textbf{34.85}} / \textcolor{red}{\textbf{0.9493}} & \textcolor{red}{\textbf{40.73}} / \textcolor{red}{\textbf{0.9824}} \\
MambaIRv2-B\cite{guo2025mambairv2} (2025) & $\times2$ & 22.9 & \textcolor{blue}{\textbf{38.65}} / \textcolor{blue}{\textbf{0.9631}} & \textcolor{blue}{\textbf{34.89}} / \textcolor{blue}{\textbf{0.9275}} & \textcolor{blue}{\textbf{32.62}} / \textcolor{blue}{\textbf{0.9053}} & \textcolor{blue}{\textbf{34.49}} / \textcolor{blue}{\textbf{0.9468}} & \textcolor{blue}{\textbf{40.42}} / \textcolor{blue}{\textbf{0.9810}} \\
MAT$^{\dagger}$\cite{xie2025mat} (2025) & $\times2$ & 9.6 & 38.61 / \textcolor{blue}{\textbf{0.9631}} & 34.53 / 0.9259 & \textcolor{blue}{\textbf{32.62}} / \textcolor{blue}{\textbf{0.9053}} & 34.31 / 0.9462 & 40.22 / 0.9808 \\
\textbf{PMS-SR (Ours)} & $\times2$ & 7.95$^{\ast}$ & 38.16 / 0.9611 & 33.95 / 0.9200 & 32.31 / 0.9002 & 33.03 / 0.9366 & 38.92 / 0.9777 \\
\textbf{PMS-SR+ (Ours)} & $\times2$ & 7.95$^{\ast}$ & 38.21 / 0.9613 & 34.06 / 0.9208 & 32.35 / 0.9005 & 33.15 / 0.9375 & 39.07 / 0.9780 \\
\midrule
EDSR\cite{lim2017enhanced} (2017) & $\times4$ & 43.1 & 32.46 / 0.8968 & 28.80 / 0.7876 & 27.71 / 0.7420 & 26.64 / 0.8033 & 31.02 / 0.9148 \\
RCAN\cite{zhang2018image} (2018) & $\times4$ & 15.6 & 32.63 / 0.9002 & 28.87 / 0.7889 & 27.77 / 0.7436 & 26.82 / 0.8087 & 31.22 / 0.9173 \\
SAN\cite{dai2019second} (2019) & $\times4$ & 15.9 & 32.64 / 0.9003 & 28.92 / 0.7888 & 27.78 / 0.7436 & 26.79 / 0.8068 & 31.18 / 0.9169 \\
DRN-L$^{\dagger}$\cite{guo2020closed} (2020) & $\times4$ & 9.8 & 32.74 / 0.9015 & 28.97 / 0.7913 & 27.83 / 0.7449 & 27.03 / 0.8131 & 31.71 / 0.9215 \\
SwinIR-M\cite{liang2021swinir} (2021) & $\times4$ & 11.9 & 32.92 / 0.9044 & 29.09 / 0.7950 & 27.92 / 0.7489 & 27.45 / 0.8254 & 32.03 / 0.9260 \\
EDT\cite{li2021efficient} (2023) & $\times4$ & 11.6 & 32.82 / 0.9031 & 29.09 / 0.7939 & 27.91 / 0.7483 & 27.46 / 0.8246 & 32.05 / 0.9254 \\
SRFormer\cite{zhou2023srformer} (2023) & $\times4$ & 10.5 & 32.93 / 0.9041 & 29.08 / 0.7953 & 27.94 / 0.7502 & 27.68 / 0.8311 & 32.21 / 0.9271 \\
HAT\cite{chen2023activating} (2023) & $\times4$ & 20.8 & 33.04 / 0.9056 & \textcolor{blue}{\textbf{29.23}} / 0.7973 & \textcolor{blue}{\textbf{28.00}} / \textcolor{blue}{\textbf{0.7517}} & \textcolor{blue}{\textbf{27.97}} / \textcolor{blue}{\textbf{0.8368}} & 32.48 / 0.9292 \\
FPAN\cite{wu2023fpan} (2023) & $\times4$ & 14.2 & 32.48 / 0.8984 & 28.78 / 0.7867 & 27.71 / 0.7412 & 26.61 / 0.8025 & 30.99 / 0.9144 \\
CLSR\cite{huang2023clsr} (2023) & $\times4$ & 15.9 & 32.76 / 0.9007 & 28.93 / 0.7903 & 27.82 / 0.7451 & 27.30 / 0.8196 & 31.76 / 0.9231 \\
SFHN\cite{wu2023sfhn} (2023) & $\times4$ & 21 & 32.74 / 0.9025 & 29.07 / 0.7932 & 27.92 / 0.7486 & 27.35 / 0.8195 & 32.05 / 0.9239 \\
CoMoNet-L\cite{huang2024como} (2024) & $\times4$ & 15.9 & 32.87 / 0.9022 & 29.08 / 0.7926 & 27.87 / 0.7461 & 27.67 / 0.8270 & 32.18 / 0.9258 \\
MambaIR\cite{guo2024mambair} (2024) & $\times4$ & 20.4 & 33.03 / 0.9046 & 29.20 / 0.7961 & 27.98 / 0.7503 & 27.68 / 0.8287 & 32.32 / 0.9272 \\
HMA\cite{chu2024hmanet} (2024) & $\times4$ & 69.9 & \textcolor{red}{\textbf{33.15}} / \textcolor{red}{\textbf{0.9060}} & \textcolor{red}{\textbf{29.32}} / \textcolor{red}{\textbf{0.7996}} & \textcolor{red}{\textbf{28.05}} / \textcolor{red}{\textbf{0.7530}} & \textcolor{red}{\textbf{28.42}} / \textcolor{red}{\textbf{0.8450}} & \textcolor{red}{\textbf{32.97}} / \textcolor{red}{\textbf{0.9320}} \\
MambaIRv2-B\cite{guo2025mambairv2} (2025) & $\times4$ & 23.1 & \textcolor{blue}{\textbf{33.14}} / \textcolor{blue}{\textbf{0.9057}} & \textcolor{blue}{\textbf{29.23}} / \textcolor{blue}{\textbf{0.7975}} & \textcolor{blue}{\textbf{28.00}} / 0.7511 & 27.89 / 0.8344 & \textcolor{blue}{\textbf{32.57}} / \textcolor{blue}{\textbf{0.9295}} \\
MAT$^{\dagger}$\cite{xie2025mat} (2025) & $\times4$ & 9.7 & 33.06 / 0.9054 & 29.17 / 0.7960 & 27.99 / 0.7514 & 27.78 / 0.8328 & 32.30 / 0.9282 \\
\textbf{PMS-SR (Ours)} & $\times4$ & 7.95$^{\ast}$ & 32.72 / 0.9012 & 29.00 / 0.7915 & 27.87 / 0.7464 & 27.25 / 0.8177 & 31.69 / 0.9222 \\
\textbf{PMS-SR+ (Ours)} & $\times4$ & 7.95$^{\ast}$ & 32.79 / 0.9020 & 29.08 / 0.7927 & 27.90 / 0.7470 & 27.38 / 0.8204 & 31.90 / 0.9237 \\
\midrule
EDSR\cite{lim2017enhanced} (2017) & $\times8$ & 45.5 & 26.96 / 0.7762 & 24.91 / 0.6420 & 24.81 / 0.5985 & 22.51 / 0.6221 & 24.69 / 0.7841 \\
D-DBPN\cite{haris2018deep} (2018) & $\times8$ & 23.2 & 27.21 / 0.7840 & 25.13 / 0.6480 & 24.88 / 0.6010 & 22.73 / 0.6312 & 25.14 / 0.7987 \\
RCAN\cite{zhang2018image} (2018) & $\times8$ & 15.7 & 27.31 / 0.7878 & 25.23 / 0.6511 & 24.98 / 0.6058 & 23.00 / 0.6452 & 25.24 / 0.8029 \\
SAN\cite{dai2019second} (2019) & $\times8$ & 16.0 & 27.22 / 0.7829 & 25.14 / 0.6476 & 24.88 / 0.6011 & 22.70 / 0.6314 & 24.85 / 0.7906 \\
DRN-L$^{\dagger}$\cite{guo2020closed} (2020) & $\times8$ & 10.0 & 27.42 / 0.7920 & 25.28 / 0.6528 & 25.01 / 0.6059 & 22.98 / 0.6434 & 25.32 / 0.8060 \\
DBPN-RES-MR64-3\cite{haris2021dbpn} (2021) & $\times8$ & 23.1 & 27.51 / 0.793 & \textcolor{blue}{\textbf{25.41}} / \textcolor{red}{\textbf{0.657}} & 25.05 / 0.607 & 23.20 / 0.652 & \textcolor{red}{\textbf{25.71}} / \textcolor{blue}{\textbf{0.813}} \\
SwinIR-M\cite{liang2021swinir} (2021) & $\times8$ & 12.1 & \textcolor{red}{\textbf{27.55}} / \textcolor{blue}{\textbf{0.7941}} & \textcolor{red}{\textbf{25.46}} / \textcolor{blue}{\textbf{0.6568}} & 25.04 / \textcolor{blue}{\textbf{0.6092}} & 23.17 / \textcolor{blue}{\textbf{0.6547}} & 25.55 / \textcolor{red}{\textbf{0.8132}} \\
RDRN\cite{panaetov2022rdrn} (2022) & $\times8$ & $\approx$36.4 & \textcolor{blue}{\textbf{27.52}} / \textcolor{red}{\textbf{0.7953}} & 25.40 / 0.6548 & \textcolor{red}{\textbf{25.10}} / 0.6084 & 23.20 / 0.6511 & \textcolor{blue}{\textbf{25.66}} / 0.8114 \\
CLSR\cite{huang2023clsr} (2023) & $\times8$ & 14.4 & 27.50 / 0.7921 & 25.35 / 0.6546 & 25.02 / 0.6070 & \textcolor{blue}{\textbf{23.22}} / \textcolor{blue}{\textbf{0.6547}} & 25.47 / 0.8102 \\
\textbf{PMS-SR (Ours)} & $\times8$ & 7.95$^{\ast}$ & 27.49 / 0.7922 & 25.37 / 0.6551 & \textcolor{blue}{\textbf{25.07}} / \textcolor{red}{\textbf{0.6094}} & \textcolor{red}{\textbf{23.36}} / \textcolor{red}{\textbf{0.6593}} & 25.58 / 0.8129 \\
\textbf{PMS-SR+ (Ours)} & $\times8$ & 7.95$^{\ast}$ & 27.54 / 0.7935 & 25.44 / 0.6568 & 25.11 / 0.6103 & 23.46 / 0.6623 & 25.75 / 0.8158 \\
\bottomrule
\end{tabular}
}
\label{tab:single_scale_comparison}
\end{table*}

\paragraph{Comparison with single-scale methods}
This comparison favors the single-input single-output (SISO) models. Each is trained for one factor and uses all of its parameters for that factor, whereas PMS-SR shares 7.95M parameters among three factors and returns all three outputs in one pass. PMS-SR is also smaller than every per-factor model in Table~\ref{tab:single_scale_comparison}. Ten of the listed methods report no $\times8$ results, and among those that report all three factors, SwinIR-M has the smallest three-model ladder, with 35.70M parameters ($4.5\times$ PMS-SR). The table is therefore a reference for how much PMS-SR gives up, or gains, by covering three factors with one model. The $^{\dagger}$ rows are our whole-image measurements of the released DRN-L and MAT \cite{xie2025mat} weights, and all other baseline values are published results; the Manga109 and $\times8$ values of EDSR and the $\times8$ values of D-DBPN follow the compilations in RCAN \cite{zhang2018image} and SAN \cite{dai2019second}, and the $\times8$ values of SwinIR-M come from its supplementary material. Our pipeline reproduces the published results of DRN-L within 0.02\,dB, those of MAT within 0.01\,dB, and those of SwinIR-M, including $\times8$, within 0.05\,dB.

At $\times8$, the hardest factor, PMS-SR outperforms EDSR, D-DBPN, RCAN, SAN, and the $\times8$-specific DRN-L \cite{guo2020closed} on all five benchmarks. On Set5 it is 0.53\,dB above the 45.5M EDSR and 0.18\,dB above RCAN, and it exceeds DRN-L by 0.38\,dB on Urban100 and by 0.26\,dB on Manga109. Against the progressive CLSR \cite{huang2023clsr}, whose $\times8$ model has 14.4M parameters, PMS-SR is 0.02--0.14\,dB higher on four benchmarks and 0.01\,dB lower on Set5. CLSR is trained on DIV2K, and our DIV2K-only control with its shortened schedule (Table~\ref{tab:multi_scale_comparison}) stays 0.03--0.34\,dB below it, so we regard the two as comparable in accuracy at $\times8$; PMS-SR reaches this level with 55\% of the parameters and returns the $\times2$ and $\times4$ images in the same pass. Three other published $\times8$ models are stronger on two or more benchmarks: SwinIR-M \cite{liang2021swinir}, DBPN-RES-MR64-3 \cite{haris2021dbpn}, and RDRN \cite{panaetov2022rdrn}, with 12.1M, 23.1M, and about 36M parameters for $\times8$ alone. Where they are ahead of PMS-SR, the margin is at most 0.13\,dB, whereas PMS-SR is at least 0.16\,dB ahead of all of them on Urban100, with the highest SSIM there as well. PMS-SR is also not significantly different from SwinIR-M on BSD100 and Manga109 in the paired test below (adjusted $p=0.17$ and $0.52$).

Because several of these margins are small, we apply the paired test of Sec.~\ref{sec:datasets} where PMS-SR loses as well as where it wins, with SwinIR-M and DRN-L re-evaluated under the same protocol. At $\times8$, PMS-SR is better than SwinIR-M on 81 of the 100 Urban100 images (median gain 0.090\,dB, adjusted $p<10^{-8}$). On the other four benchmarks, the differences are not significant (adjusted $p>0.16$). Against DRN-L the advantage is broader: PMS-SR is better on all 100 Urban100 images (median gain 0.249\,dB), on 102 of the 109 Manga109 images (0.191\,dB), and on 85 of the 100 BSD100 images (adjusted $p<10^{-9}$ in all three cases). At $\times2$ and $\times4$, SwinIR-M stays ahead on BSD100, Urban100, and Manga109 (adjusted $p<10^{-9}$).

\paragraph{Gap to the single-scale models across factors}
The gap to SwinIR-M in Table~\ref{tab:single_scale_comparison} shrinks with the factor: averaged over the five benchmarks, it is $-0.55$\,dB at $\times2$, $-0.18$\,dB at $\times4$, and $+0.02$\,dB at $\times8$. It never widens with the factor on any benchmark and is positive at $\times8$ on BSD100, Urban100, and Manga109 ($+0.03$, $+0.19$, and $+0.03$\,dB), significantly so on Urban100. The same pattern holds against CLSR, which also reports all three factors: the average differences are $-0.41$, $-0.01$, and $+0.06$\,dB. At $\times4$, PMS-SR is also ahead of the DIV2K-trained EDSR, RCAN, and SAN on all five benchmarks. Among the recent models listed only at $\times2$ and $\times4$, FPAN is 0.16--0.70\,dB below PMS-SR at $\times4$ and within 0.31\,dB of it at $\times2$, whereas SFHN, CoMoNet-L, and MAT, with a separate 9.6M--21M model per factor, lead by up to 0.36, 0.49, and 0.61\,dB at $\times4$ and by 0.13--1.30\,dB at $\times2$. This trend is consistent with the design: the objective gives the largest weight to the hardest factor \eqref{eq:total}, the cascaded anchors keep the shared backbone from being spent on the factor that bicubic interpolation already serves well, and the IMR gains are larger at $\times4$ and $\times8$ than at $\times2$ (Table~\ref{tab:ablation_imr}). A system that must deliver all three renditions thus gives up 0.2--1.0\,dB at $\times2$ against an 11.8M single-scale model and obtains $\times8$ results on par with the strongest single-scale $\times8$ models, with 22\% of the parameters of the three-model ladder.

\paragraph{Self-ensemble}
At eight times the inference cost, PMS-SR+ gains $+0.03$ to $+0.21$\,dB in all fifteen benchmark--factor combinations. Against published self-ensembles at $\times8$, it is ahead of RCAN+ \cite{zhang2018image} on all five benchmarks, of SwinIR+ \cite{liang2021swinir} on BSD100, Urban100, and Manga109, and of RDRN+ \cite{panaetov2022rdrn} only on Urban100.

\subsection{Qualitative Comparison}
\label{sec:qual}
\begin{figure*}[!t]
    \centering
    \includegraphics[width=0.73\textwidth,trim=0 28.8bp 0 0,clip]{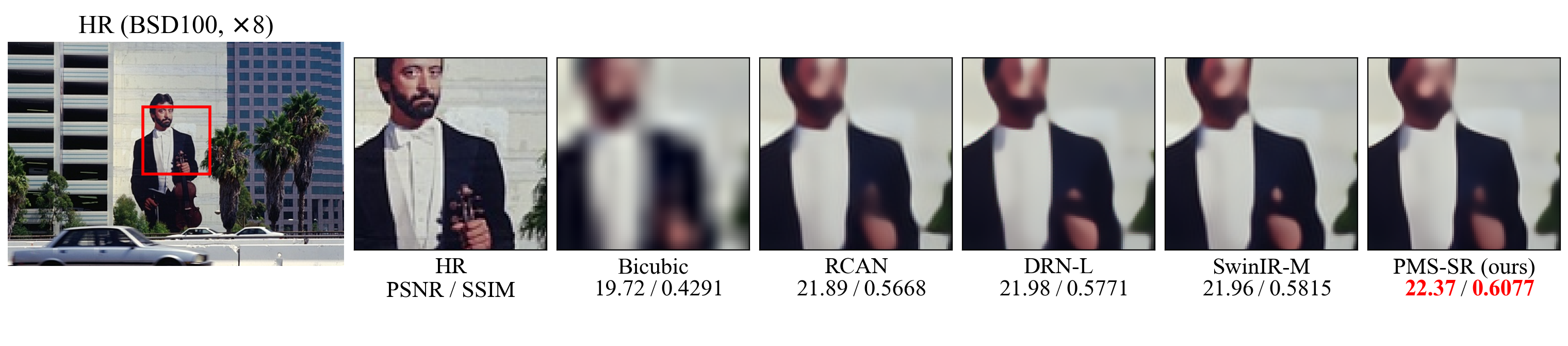}\\[-1pt]
    {\footnotesize (a)}\\[2pt]
    \includegraphics[width=0.73\textwidth,trim=0 28.8bp 0 0,clip]{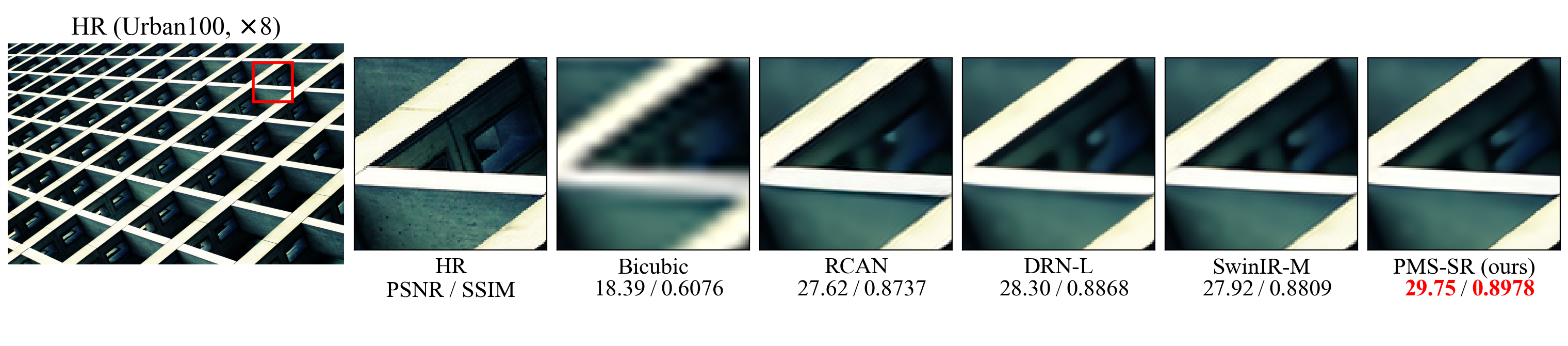}\\[-1pt]
    {\footnotesize (b)}\\[2pt]
    \includegraphics[width=0.73\textwidth,trim=0 28.8bp 0 0,clip]{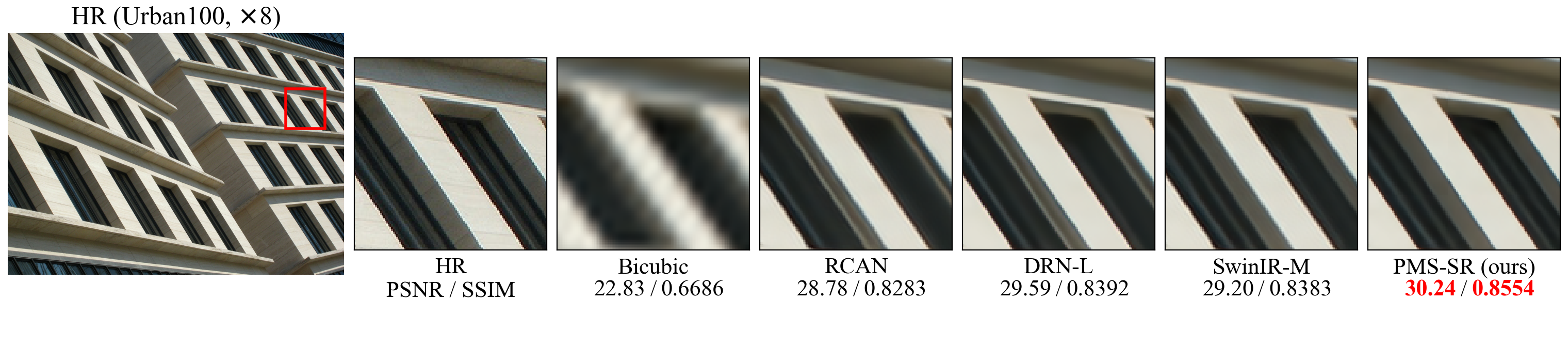}\\[-1pt]
    {\footnotesize (c)}\\[2pt]
    \includegraphics[width=0.73\textwidth,trim=0 28.8bp 0 0,clip]{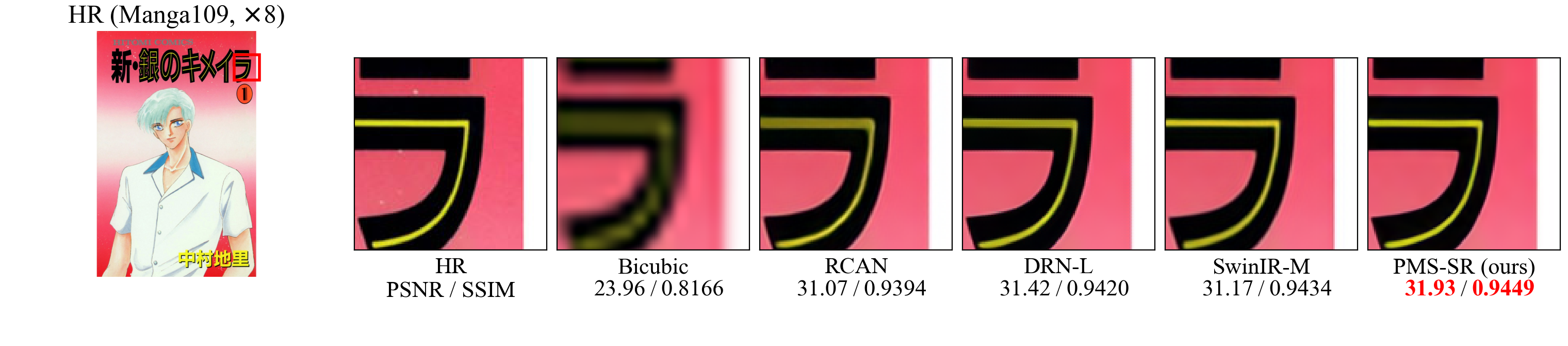}\\[-1pt]
    {\footnotesize (d)}
    \caption{Visual comparison at $\times8$: (a) BSD100 violinist, (b) Urban100 img\_043, (c) Urban100 img\_025, and (d) Manga109 \emph{GinNoChimera}. Full-image PSNR (dB)\,/\,SSIM is given below each panel, and the best value of each metric is shown in \textcolor{red}{\textbf{bold red}}.}
    \label{fig:cmp1}
\end{figure*}

\begin{figure*}[t]
    \centering
    \includegraphics[width=0.73\textwidth,trim=0 28.8bp 0 0,clip]{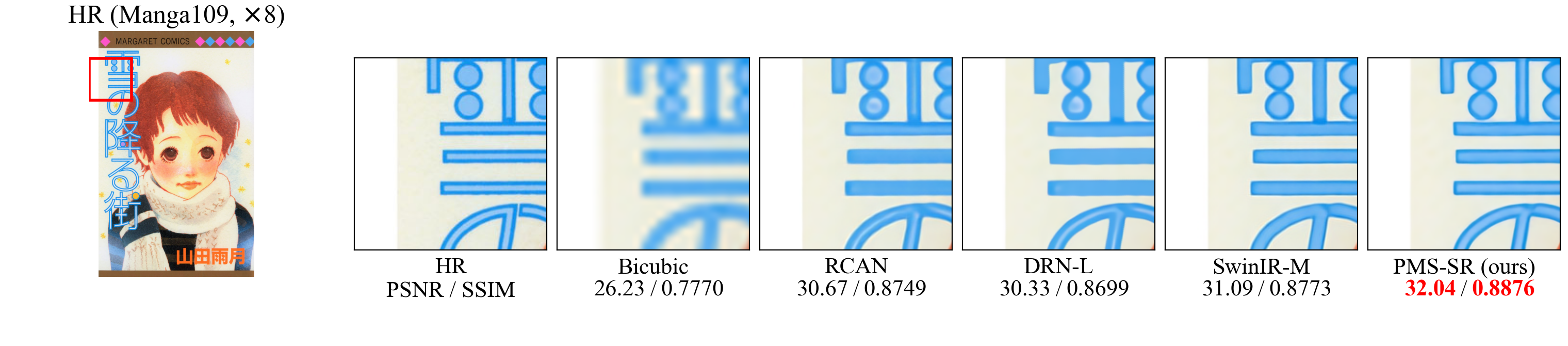}\\[-1pt]
    {\footnotesize (a)}\\[2pt]
    \includegraphics[width=0.73\textwidth,trim=0 28.8bp 0 0,clip]{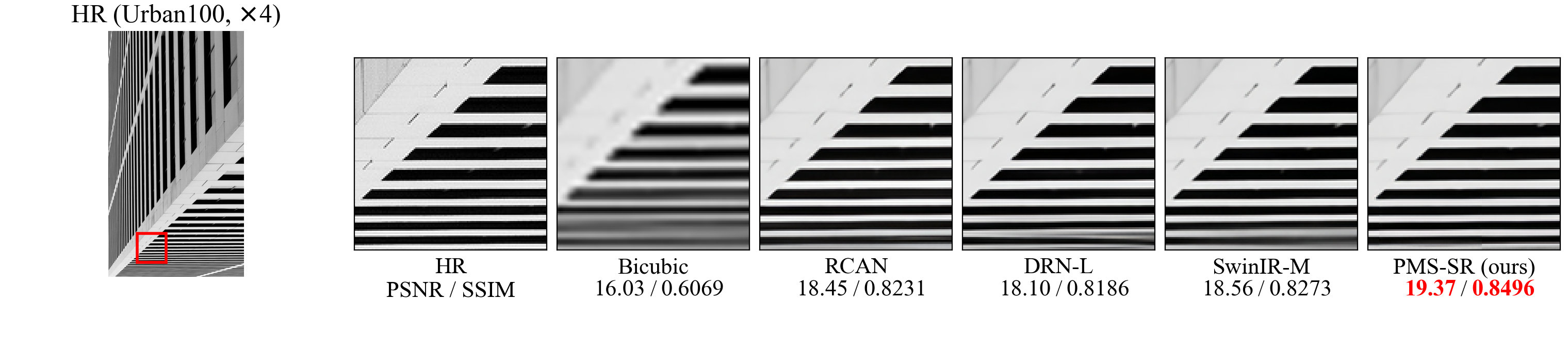}\\[-1pt]
    {\footnotesize (b)}\\[2pt]
    \includegraphics[width=0.73\textwidth,trim=0 28.8bp 0 0,clip]{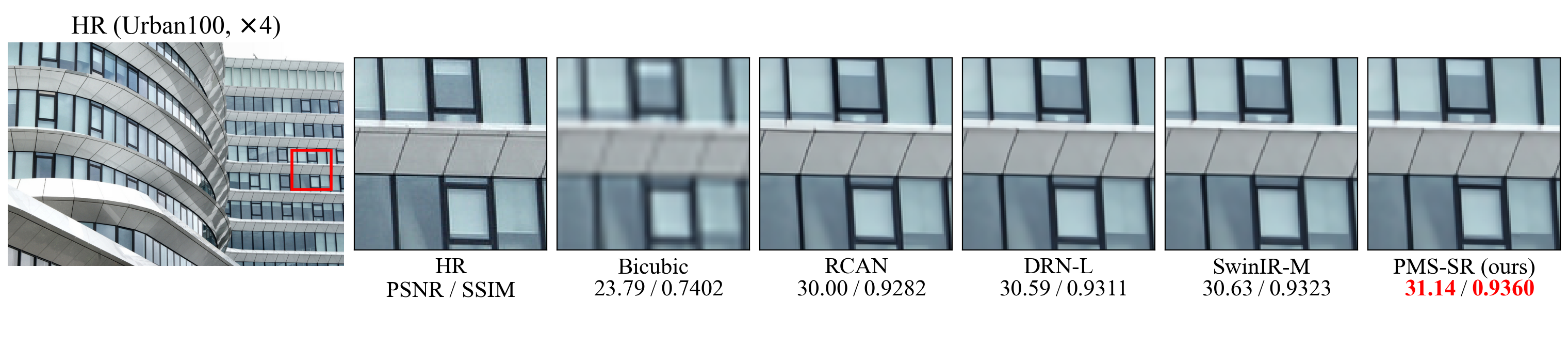}\\[-1pt]
    {\footnotesize (c)}
    \caption{Visual comparison at (a) $\times8$ on Manga109 (\emph{YukiNoFuruMachi}), (b) $\times4$ on Urban100 img\_011, and (c) $\times4$ on Urban100 img\_052. Protocol and highlighting are the same as in Fig.~\ref{fig:cmp1}.}
    \label{fig:cmp2}
\end{figure*}

Fig.~\ref{fig:cmp1} compares PMS-SR at $\times8$ with bicubic interpolation, RCAN \cite{zhang2018image}, the $\times8$-specific DRN-L \cite{guo2020closed}, and the released SwinIR-M $\times8$ model \cite{liang2021swinir}. All methods receive the same LR input, and we ran the baselines with the authors' weights. The examples are images on which PMS-SR leads all three networks in both PSNR and SSIM; at $\times8$ this holds for 31, 60, and 32 images of BSD100, Urban100, and Manga109, and at $\times4$ for 9 images, whereas at $\times2$ the single-scale Transformers lead on almost every image (Sec.~\ref{sec:quant}), so no $\times2$ example is shown. At $\times8$, PMS-SR is ahead of the best baseline by 0.39\,dB on the BSD100 violinist, 1.45 and 0.65\,dB on the two Urban100 fa\c{c}ades, and 0.51\,dB on the Manga109 title in Fig.~\ref{fig:cmp1}. On the Manga109 page in Fig.~\ref{fig:cmp2}(a), it is 0.95--1.71\,dB better than all three baselines and keeps the title strokes separate where the baselines merge or thicken them. In the two $\times4$ cases of Fig.~\ref{fig:cmp2}(b) and (c), it is 0.81 and 0.51\,dB ahead of SwinIR-M. Each PMS-SR result comes from a pass that also produced the other two renditions.

\paragraph{Perceptual quality}
Under the same protocol, the $\times8$ LPIPS of PMS-SR on Set5, Set14, BSD100, Urban100, and Manga109 is 0.284/0.428/0.534/0.361/0.214. It is within 0.011 of SwinIR-M on every benchmark, lower than RCAN on all five, and lower than DRN-L on all except Set5. Like these baselines, PMS-SR is trained with a distortion-oriented objective, so its gains lie in fidelity and cross-factor coverage rather than in perceptual quality.

\subsection{Ablation Studies}
\label{sec:ablation}

\paragraph{Effectiveness of the Inter-Scale Mutual Refinement Stage}
Evaluating the Stage-1 outputs $\hat{I}^{(1)}_{s}$ of the same trained model removes the IMR module at test time (Table~\ref{tab:ablation_imr}, run on Set5, Set14, BSD100, and Urban100). IMR improves PSNR on all four benchmarks at every scale, with the largest gains on Urban100 ($+0.08$ to $+0.10$\,dB at all three scales and $+0.0044$ SSIM at $\times8$), where repeated structures benefit most from cross-scale alignment. In the paired test, the full model is better on 87--98 of the 100 images of BSD100 and Urban100 at every scale (adjusted $p<10^{-12}$). It is also better on 11--12 of the 14 Set14 images (adjusted $p\le0.022$), whereas the differences on the five Set5 images are not significant. Because this is a test-time ablation of jointly trained weights, the gains measure the contribution of Stage~2 within the trained model; axis~(a) below measures the effect of training without it.

\begin{table}[t]
\centering
\small
\caption{Test-time ablation of the IMR module: Stage-1 outputs (w/o IMR) vs.\ the full model, identical trained weights (PSNR\,/\,SSIM, Y channel; better in \textcolor{red}{red}, ties unmarked).}
\label{tab:ablation_imr}
\renewcommand{\arraystretch}{1.0}
\setlength{\tabcolsep}{3pt}
\resizebox{\columnwidth}{!}{%
\begin{tabular}{llcccc}
\toprule
Scale & Variant & Set5 & Set14 & BSD100 & Urban100 \\
\midrule
\multirow{2}{*}{$\times2$} & w/o IMR & 38.14 / 0.9611 & 33.93 / 0.9198 & 32.30 / 0.9001 & 32.95 / 0.9361 \\
 & Full & \textcolor{red}{\textbf{38.16}} / 0.9611 & \textcolor{red}{\textbf{33.95}} / \textcolor{red}{\textbf{0.9200}} & \textcolor{red}{\textbf{32.31}} / \textcolor{red}{\textbf{0.9002}} & \textcolor{red}{\textbf{33.03}} / \textcolor{red}{\textbf{0.9366}} \\
\midrule
\multirow{2}{*}{$\times4$} & w/o IMR & 32.64 / 0.9006 & 28.96 / 0.7906 & 27.84 / 0.7458 & 27.15 / 0.8154 \\
 & Full & \textcolor{red}{\textbf{32.72}} / \textcolor{red}{\textbf{0.9012}} & \textcolor{red}{\textbf{29.00}} / \textcolor{red}{\textbf{0.7915}} & \textcolor{red}{\textbf{27.87}} / \textcolor{red}{\textbf{0.7464}} & \textcolor{red}{\textbf{27.25}} / \textcolor{red}{\textbf{0.8177}} \\
\midrule
\multirow{2}{*}{$\times8$} & w/o IMR & 27.43 / 0.7900 & 25.34 / 0.6535 & 25.05 / 0.6083 & 23.27 / 0.6549 \\
 & Full & \textcolor{red}{\textbf{27.49}} / \textcolor{red}{\textbf{0.7922}} & \textcolor{red}{\textbf{25.37}} / \textcolor{red}{\textbf{0.6551}} & \textcolor{red}{\textbf{25.07}} / \textcolor{red}{\textbf{0.6094}} & \textcolor{red}{\textbf{23.36}} / \textcolor{red}{\textbf{0.6593}} \\
\bottomrule
\end{tabular}}
\end{table}

\begin{figure*}[!b]
    \centering
    \includegraphics[width=0.74\textwidth]{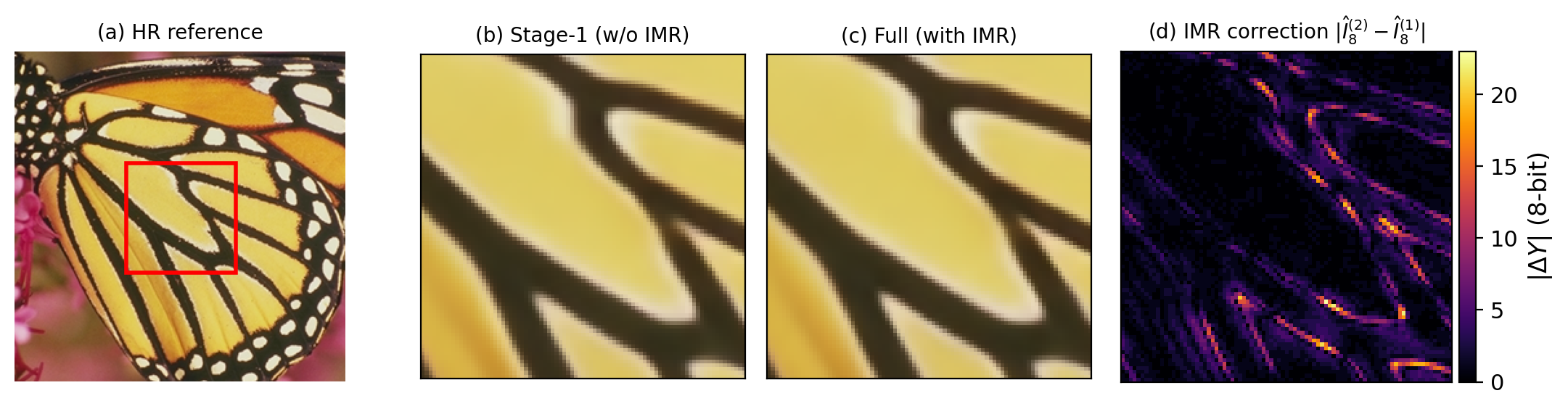}
    \caption{Effect of IMR at $\times8$: (a) HR image with the zoomed region, (b) Stage-1 output (without IMR), (c) full model (with IMR), and (d) the correction $|\hat{I}^{(2)}_{8}-\hat{I}^{(1)}_{8}|$ on the Y channel, which concentrates along edges and texture transitions.}
    \label{fig:imr_viz}
\end{figure*}

Fig.~\ref{fig:imr_viz} shows that the Stage-2 correction is sparse: it concentrates on edges and texture transitions that the cascade does not fully resolve and leaves flat regions almost unchanged.

\paragraph{Further ablation axes}
Separating the individual mechanisms requires retraining, so we train five variants from scratch with a shortened but otherwise identical 50k-iteration schedule, each using the checkpoint selected as in Sec.~\ref{sec:impl} (Table~\ref{tab:ablation_axes}): (a)~no Stage~2, (b)~no channel attention branch, (c)~a direct $\mathrm{bic}_{\uparrow s}(I_{\mathrm{LR}})$ anchor instead of the cascade, (d)~no anchor detachment, and (e)~concatenation instead of deformable alignment. They are comparable with one another but not with the 300k model. Two mechanisms have a clear effect at this budget. Replacing the cascaded anchor (c) causes large losses exactly where the cascade operates: $-0.99$\,dB at $\times4$ and $-2.38$\,dB at $\times8$ on BSD100 ($-2.61$ and $-4.46$\,dB on Set5), whereas $\times2$, whose anchor is the same in both designs, is unchanged, as the design predicts. Concatenation (e) costs 0.24--0.29\,dB on BSD100 at every scale. Variants (a), (b), and (d) stay within $\pm0.02$\,dB of the reference on BSD100 at this budget; Stage~2 nevertheless gives systematic gains with the full schedule (Table~\ref{tab:ablation_imr}), and detachment is kept because ProSR reports instability for anchors built on predicted outputs \cite{wang2018prosr}.

\begin{table}[t]
\centering
\small
\caption{Further ablation axes, trained from scratch for 50k iterations (Y-PSNR, dB). Drops of at least $0.2$\,dB vs.\ the 50k reference in \textbf{bold}; all others are within $\pm0.1$\,dB.}
\label{tab:ablation_axes}
\renewcommand{\arraystretch}{1.0}
\setlength{\tabcolsep}{4pt}
\resizebox{\columnwidth}{!}{%
\begin{tabular}{lcccccc}
\toprule
 & \multicolumn{3}{c}{BSD100} & \multicolumn{3}{c}{Set5} \\
\cmidrule(lr){2-4}\cmidrule(lr){5-7}
Variant & $\times2$ & $\times4$ & $\times8$ & $\times2$ & $\times4$ & $\times8$ \\
\midrule
Full model (50k reference) & 32.19 & 27.75 & 24.98 & 38.00 & 32.47 & 27.26 \\
\midrule
(a) w/o Stage~2 (IMR), from scratch & 32.20 & 27.74 & 24.96 & 38.01 & 32.46 & 27.24 \\
(b) w/o channel attention branch & 32.19 & 27.73 & 24.96 & 38.01 & 32.47 & 27.23 \\
(c) direct $\mathrm{bic}_{\uparrow s}$ anchor & 32.21 & \textbf{26.76} & \textbf{22.60} & 38.03 & \textbf{29.86} & \textbf{22.80} \\
(d) cascade without detachment & 32.21 & 27.75 & 24.99 & 38.01 & 32.55 & 27.29 \\
(e) concat instead of deformable align. & \textbf{31.95} & \textbf{27.46} & \textbf{24.73} & \textbf{37.65} & \textbf{31.91} & \textbf{26.78} \\
\bottomrule
\end{tabular}}
\end{table}

\begin{table}[t]
\centering
\small
\caption{Cost of producing the full $\times2/{\times}4/{\times}8$ ladder for one input. FLOPs are counted by thop for one $64{\times}64$ LR input; runtimes are averaged over the 100 BSD100 $\times8$ LR inputs (about $60{\times}40$) on an RTX~4090. Tiled: $48{\times}48$ tiles with overlap 16.}
\label{tab:complexity}
\renewcommand{\arraystretch}{1.0}
\setlength{\tabcolsep}{4pt}
\resizebox{\columnwidth}{!}{%
\begin{tabular}{lccc}
\toprule
Method (ladder) & Params & FLOPs & Runtime \\
\midrule
SwinIR-M $\times2$ \cite{liang2021swinir} & 11.75M & 48.05G & 30.3\,ms \\
SwinIR-M $\times4$ & 11.90M & 50.55G & 31.5\,ms \\
SwinIR-M $\times8$ & 12.05M & 60.55G & 32.6\,ms \\
SwinIR-M ladder (3 models, 3 passes) & 35.70M & 159.15G & 94.4\,ms \\
Derived ladder (SwinIR-M $\times8$ $+$ bicubic$\downarrow$) & 12.05M & 60.55G & 32.6\,ms \\
\midrule
\textbf{PMS-SR} (1 model, 1 pass, whole image) & 7.95M & 142.17G & 32.8\,ms \\
\textbf{PMS-SR} (1 model, 1 pass, tiled) & 7.95M & 319.88G & 47.1\,ms \\
\bottomrule
\end{tabular}}
\end{table}

\subsection{Efficiency and Cross-Scale Consistency}
\label{sec:eff}

\begin{figure}[!t]
\centering
\includegraphics[width=0.90\columnwidth]{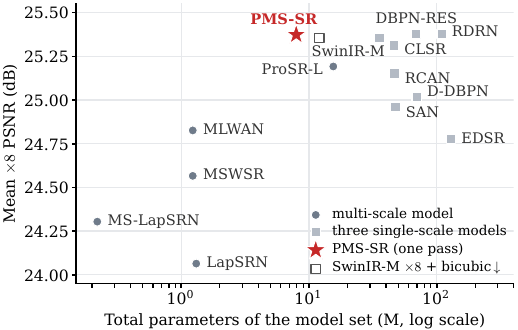}
\caption{Mean $\times8$ PSNR over the five benchmarks vs.\ the total parameters of the models that deliver $\times2$, $\times4$, and $\times8$ (Tables~\ref{tab:multi_scale_comparison}, \ref{tab:single_scale_comparison}, and~\ref{tab:complexity}). Single-scale methods count as three models; where one size is reported, it is tripled. DRN-L has no $\times2$ model. The hollow marker is the derived ladder.}
\label{fig:ladder}
\end{figure}

\paragraph{Complexity}
Table~\ref{tab:complexity} compares the cost of producing all three renditions of one input, measured in parameters, floating-point operations (FLOPs), and runtime. SwinIR-M is the strongest competitor with released weights for all three factors. Whole-image inference, whose accuracy is close to that of the tiled protocol (Sec.~\ref{sec:impl}), needs 142.17\,GFLOPs, fewer than the 159.15\,GFLOPs of the SwinIR-M ladder, because attention runs only on the LR grid and the cross-scale exchange for $\times8$ runs on the $\times4$ grid. The tiled mode processes overlapping tiles and needs 2.25 times as many FLOPs. Fig.~\ref{fig:ladder} summarizes the trade-off at $\times8$: PMS-SR reaches the mean $\times8$ PSNR of SwinIR-M, CLSR, DBPN-RES-MR64-3, and RDRN (25.37 vs.\ 25.31--25.38\,dB) with at most 22\% of the parameters of their three-model ladders, and all smaller models in the figure are at least 0.54\,dB lower.

\begin{table}[!t]
\centering
\small
\caption{Single-input deployment: each system receives the same $\times8$ LR input and delivers all three renditions (whole-image inference). Italic rows are degenerate controls; the best fidelity per column (controls excluded) is in \textcolor{red}{red}.}
\label{tab:consistency}
\renewcommand{\arraystretch}{0.92}
\setlength{\tabcolsep}{2.6pt}
\scriptsize
\begin{tabular}{@{}llccccc@{}}
\toprule
 & & \multicolumn{2}{c}{Agreement (dB) $\uparrow$} & \multicolumn{3}{c}{Fidelity (dB) $\uparrow$} \\
\cmidrule(lr){3-4}\cmidrule(lr){5-7}
Benchmark & System & $\times8\!\rightarrow\!\times4$ & $\times4\!\rightarrow\!\times2$ & $\times2$ & $\times4$ & $\times8$ \\
\midrule
\multirow{4}{*}{Set5}
 & \textit{Bicubic ladder} (control) & 62.48 & 55.16 & 27.72 & 25.36 & 24.40 \\
 & \textit{Derived ladder} (control) & \textit{$\infty$} & \textit{$\infty$} & 33.24 & 29.32 & 27.59 \\
 & SwinIR-M (3 models) & 40.52 & 42.54 & 32.89 & \textcolor{red}{\textbf{29.33}} & \textcolor{red}{\textbf{27.59}} \\
 & \textbf{PMS-SR} (1 model, 1 pass) & \textcolor{red}{\textbf{52.36}} & \textcolor{red}{\textbf{50.21}} & \textcolor{red}{\textbf{33.09}} & 29.19 & 27.49 \\
\midrule
\multirow{4}{*}{Set14}
 & \textit{Bicubic ladder} (control) & 62.10 & 55.32 & 27.73 & 24.74 & 23.19 \\
 & \textit{Derived ladder} (control) & \textit{$\infty$} & \textit{$\infty$} & 32.31 & 27.77 & 25.46 \\
 & SwinIR-M (3 models) & 40.14 & 42.16 & 32.00 & \textcolor{red}{\textbf{27.84}} & \textcolor{red}{\textbf{25.46}} \\
 & \textbf{PMS-SR} (1 model, 1 pass) & \textcolor{red}{\textbf{50.57}} & \textcolor{red}{\textbf{49.18}} & \textcolor{red}{\textbf{32.24}} & 27.68 & 25.37 \\
\midrule
\multirow{4}{*}{BSD100}
 & \textit{Bicubic ladder} (control) & 62.91 & 56.89 & 29.03 & 25.65 & 23.67 \\
 & \textit{Derived ladder} (control) & \textit{$\infty$} & \textit{$\infty$} & 32.42 & 27.71 & 25.05 \\
 & SwinIR-M (3 models) & 42.54 & 44.25 & 32.34 & \textcolor{red}{\textbf{27.79}} & 25.05 \\
 & \textbf{PMS-SR} (1 model, 1 pass) & \textcolor{red}{\textbf{51.48}} & \textcolor{red}{\textbf{50.91}} & \textcolor{red}{\textbf{32.58}} & 27.77 & \textcolor{red}{\textbf{25.07}} \\
\midrule
\multirow{4}{*}{Urban100}
 & \textit{Bicubic ladder} (control) & 62.40 & 54.75 & 25.94 & 22.62 & 20.74 \\
 & \textit{Derived ladder} (control) & \textit{$\infty$} & \textit{$\infty$} & 32.15 & 26.38 & 23.19 \\
 & SwinIR-M (3 models) & 34.53 & 38.30 & 31.75 & \textcolor{red}{\textbf{26.64}} & 23.19 \\
 & \textbf{PMS-SR} (1 model, 1 pass) & \textcolor{red}{\textbf{46.91}} & \textcolor{red}{\textbf{46.98}} & \textcolor{red}{\textbf{32.32}} & 26.57 & \textcolor{red}{\textbf{23.35}} \\
\midrule
\multirow{4}{*}{Manga109}
 & \textit{Bicubic ladder} (control) & 62.83 & 54.46 & 25.46 & 22.50 & 21.46 \\
 & \textit{Derived ladder} (control) & \textit{$\infty$} & \textit{$\infty$} & 32.24 & 27.45 & 25.55 \\
 & SwinIR-M (3 models) & 35.80 & 38.45 & 31.65 & \textcolor{red}{\textbf{27.67}} & \textcolor{red}{\textbf{25.55}} \\
 & \textbf{PMS-SR} (1 model, 1 pass) & \textcolor{red}{\textbf{45.64}} & \textcolor{red}{\textbf{45.54}} & \textcolor{red}{\textbf{32.46}} & 27.55 & 25.54 \\
\bottomrule
\end{tabular}
\end{table}

\paragraph{Cost of the ladder}
On an otherwise idle RTX~4090, one whole-image PMS-SR pass delivers all three renditions in 32.8\,ms, against 94.4\,ms for the SwinIR-M ladder. The $2.9\times$ speed-up is much larger than the $1.12\times$ FLOPs ratio because the ladder pays the fixed per-network overhead three times; for larger inputs, the speed-up is expected to approach the FLOPs ratio. The tiled mode takes 47.1\,ms and is still $2.0\times$ faster. The derived ladder is the cheapest option (32.6\,ms), but all of its renditions are tied to the $\times8$ output.

\paragraph{Cross-scale consistency}
A SIMO model delivers several renditions of one input, so a property that per-factor evaluation never measures becomes relevant: whether the renditions agree. Table~\ref{tab:consistency} measures this for a single LR input as the Y-PSNR between the downsampled higher-factor output and the delivered lower-factor output. PMS-SR reaches 45.6--52.4\,dB ($\times8\!\rightarrow\!\times4$) and 45.5--50.9\,dB ($\times4\!\rightarrow\!\times2$), against 34.5--42.5\,dB and 38.3--44.3\,dB for three independent SwinIR-M models, an advantage of 6.7--12.4\,dB that is largest on Urban100. Two independent models of the more recent MAT \cite{xie2025mat} behave likewise: on the same inputs, their $\times2$ and $\times4$ outputs agree at 38.9--44.0\,dB, which is 6.7--8.0\,dB below PMS-SR. The MAT renditions are up to 0.41\,dB more accurate at $\times4$, and those of PMS-SR are 0.22--0.53\,dB more accurate at $\times2$ on four benchmarks (0.03\,dB less on Urban100).

Agreement alone is not evidence of quality. A bicubic ladder reaches about 62\,dB at $\times8\!\rightarrow\!\times4$ because it adds no detail, and a derived ladder (a $\times8$ model followed by downsampling) is consistent by construction, so agreement must be read together with fidelity. In the fidelity columns, PMS-SR gives the most accurate $\times2$ renditions of the listed non-control systems on all benchmarks, and on the two 100-image benchmarks its $\times2$ and $\times8$ renditions are the most accurate in the table (e.g., 32.58 and 25.07\,dB on BSD100). PMS-SR is also more accurate than the derived ladder at every factor on BSD100 and Urban100 (by up to 0.19\,dB), whereas the derived ladder, whose $\times2$ and $\times4$ renditions are only downsampled $\times8$ outputs, is 0.07--0.15\,dB ahead on the small Set5 and Set14. Three independent models have competitive fidelity but the weakest agreement.

\section{Discussion}
\label{sec:discussion}

Three design decisions keep $\times8$ affordable in one SIMO pass: deep feature extraction is confined to the LR grid with a narrowing channel pyramid, detached cascaded anchors limit each head to the band the previous scale lacks (the most important axis in Table~\ref{tab:ablation_axes}), and IMR couples the scales in both directions, with larger gains at $\times4$ and $\times8$ than at $\times2$ (Table~\ref{tab:ablation_imr}). Our training also uses only about 30\% of the sample budget of SwinIR, so further gains may be possible.

For video systems, the single-pass design suits services that keep several renditions of one source, such as the resolution ladders of adaptive streaming \cite{menon2024jtps,amirpour2025deepstream}. An edge node that upscales a low-bitrate stream can produce the whole ladder in one pass instead of running three SwinIR-M-class models with $4.5\times$ the parameters. Because the renditions are mutually consistent (Sec.~\ref{sec:eff}), switching between factors changes the viewed content much less than with independent models.

\subsection{Limitations}
\label{sec:limitations}
Three limitations should be noted. First, the accuracy ranking depends on the factor: PMS-SR is on par with the strongest single-scale models at $\times8$ and competitive at $\times4$, whereas at $\times2$ single-scale Transformers, each larger than PMS-SR, remain 0.2--1.8\,dB ahead (Table~\ref{tab:single_scale_comparison}). Our objective gives the largest weight to the hardest factor \eqref{eq:total}, and the $\times2$ and $\times4$ heads are trained against bicubically downsampled targets (Sec.~\ref{sec:loss}), which may also contribute; a deployment that prioritizes $\times2$ would need to rebalance these weights. Second, the sub-pixel pyramid restricts the factors to powers of two. Non-dyadic factors require an extra branch or resampling, and arbitrary-scale operation \cite{chen2021learning,zhang2026cgssr} is outside our scope. Third, all experiments use single images with the bicubic degradation. Real-world degradations \cite{zhang2021designing,zheng2026realarbvsr} and temporal consistency are not modeled, and cross-scale consistency (Table~\ref{tab:consistency}) does not imply cross-frame consistency; extending the cascaded anchors and IMR along the time axis is a natural next step.

\section{Conclusion}
\label{sec:conclusion}

We introduced PMS-SR, a progressive Transformer-based SIMO framework that reconstructs $\times2$, $\times4$, and $\times8$ images in a single forward pass. Its CSWin backbone is deliberately standard and confined to the LR grid. The contribution lies in how the three renditions are coupled: cascaded anchors start the optimization exactly from the bicubic cascade and limit each head to the band the previous scale lacks, and inter-scale mutual refinement links the scales in both directions, with the $\times8$ refinement computed on the $\times4$ grid. With 7.95M parameters, PMS-SR has the highest PSNR among the compared multi-scale methods in every $\times4$ and $\times8$ entry. At $\times8$, it outperforms the $\times8$-specific DRN-L on all five benchmarks, leads all compared methods on Urban100, and stays within 0.13\,dB of the strongest $\times8$-specific models elsewhere. PMS-SR is smaller than any single-scale model in Table~\ref{tab:single_scale_comparison} and, with 22\% of the parameters of three SwinIR-M models, delivers all three renditions in one pass, $2.9\times$ faster on BSD100 inputs. Future work includes curriculum-style scale scheduling \cite{wang2018prosr}, support for arbitrary factors, and application of both mechanisms to video SR and joint restoration.

\section*{Acknowledgment}
The authors would like to thank the National Science and Technology Council, Taiwan, for supporting this work under Grant NSTC 114-2221-E-005-040-MY2.

\begingroup
\setlength{\parskip}{0pt}
\setlength{\parindent}{1em}
\bibliographystyle{IEEEtran}
\bibliography{reference}
\endgroup

\raggedbottom

\makeatletter
\def\@IEEEBIOskipN{1\baselineskip}
\makeatother

\begin{IEEEbiography}[{\includegraphics[width=1in,height=1.25in,clip,keepaspectratio]{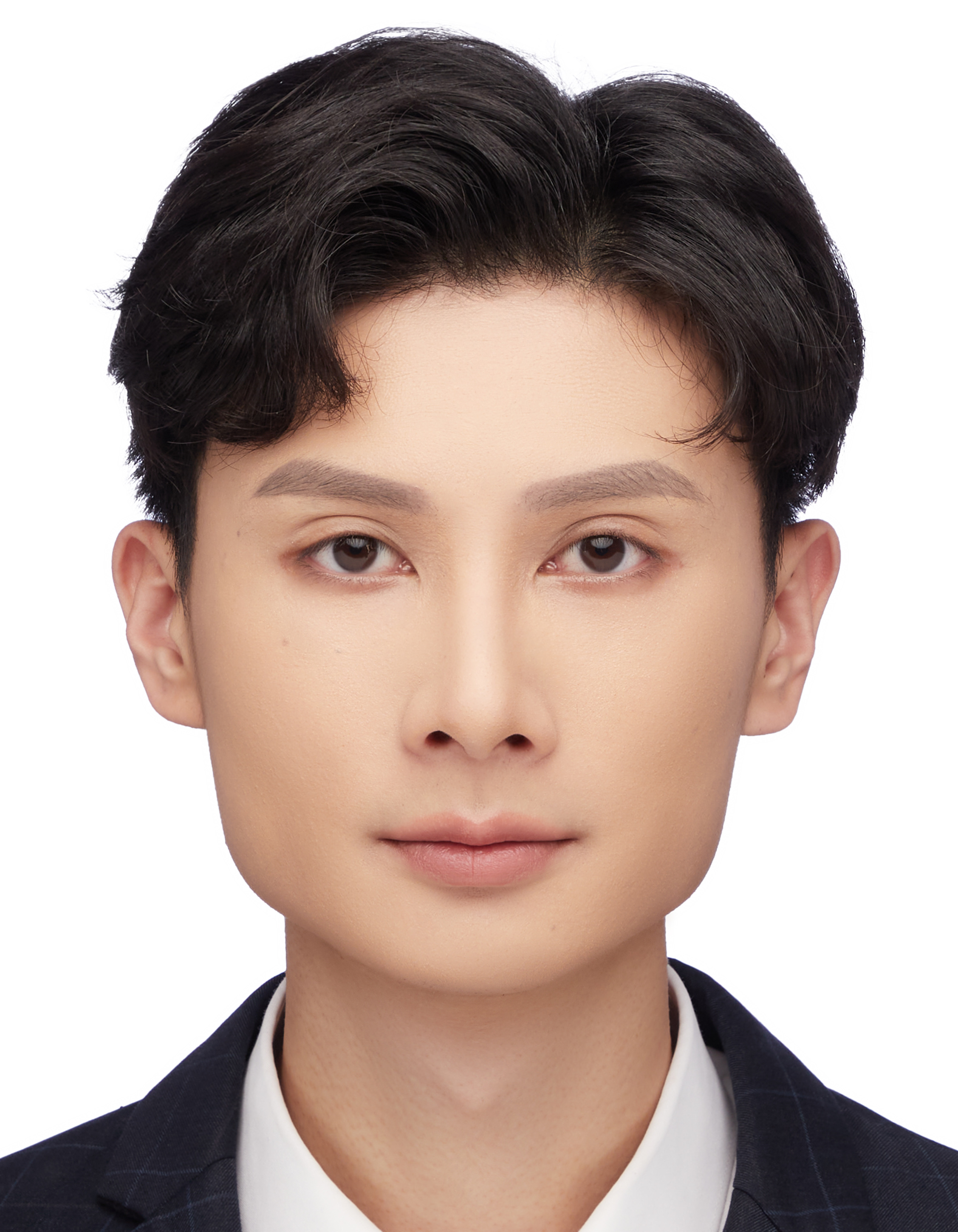}}]{Chun-An Lin}
(Student Member, IEEE) is currently pursuing the Ph.D. degree with the Department of Electrical Engineering, National Chung Hsing University, Taiwan, and a Lecturer with the College of Computing and Informatics, Providence University, Taiwan. He received his M.S. degree from the Department of Multimedia Design, National Taichung University of Science and Technology, Taiwan, in 2021 and his B.S. degree from the Department of Computer Science \& Communication Engineering, Providence University, Taiwan, in 2019. His research focuses on Machine Learning, Deep Learning, Image Super-Resolution, and Medical Image Analysis.
\end{IEEEbiography}

\begin{IEEEbiography}[{\includegraphics[width=1in,height=1.25in,clip,keepaspectratio]{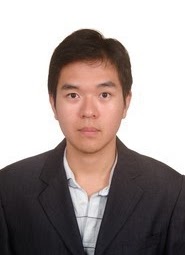}}]{Yu-Chia Chang}
received the M.S. degree from the Department of Electrical Engineering, National Chung Hsing University, Taiwan, in 2025, and his B.S. degree from the Department of Electrical Engineering, Yuan Ze University, Taiwan, in 2018. His research focuses on Image Super-Resolution, Object Detection, Image Processing, Machine Learning, Deep Learning, Reinforcement Learning, and Feature Engineering.
\end{IEEEbiography}

\begin{IEEEbiography}[{\includegraphics[width=1in,height=1.25in,clip,keepaspectratio]{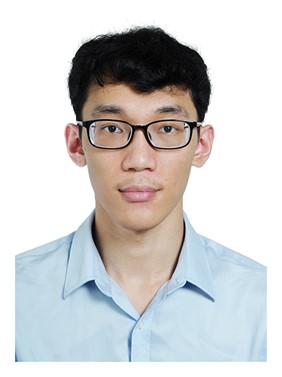}}]{Chi-Mao Fan}
is currently an AI R\&D Engineer in the Digital Image Technology Department (DIT) at ASUS, Taipei, Taiwan. He received his M.S. degree in Electrical Engineering from National Chung Hsing University, Taiwan, in 2023 and his B.S. degree in Electrical Engineering from Yuan Ze University, Taiwan, in 2021. His research interests include Computer Vision, Image Restoration, Image Denoising, Oriented Object Detection, and Retrieval-Augmented Generation, with a focus on bridging academic research and industrial applications.
\end{IEEEbiography}

\begin{IEEEbiography}[{\includegraphics[width=1in,height=1.25in,clip,keepaspectratio]{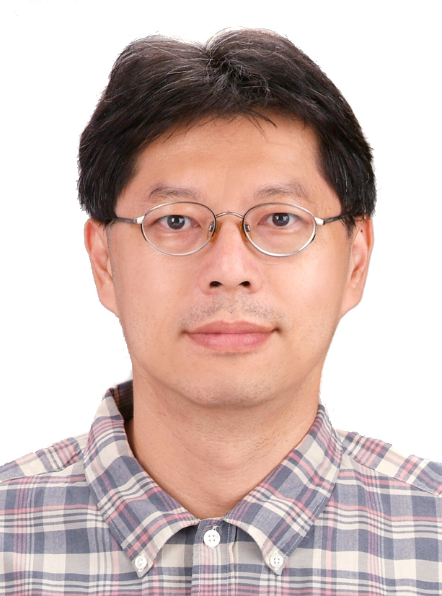}}]{Yen-Chieh Ouyang}
(Life Member, IEEE) received the Bachelor of Science degree in electrical engineering from Feng Chia University, Taiwan, in 1981, and the Master of Science and Doctor of Philosophy degrees in electrical engineering from The University of Memphis, Memphis, TN, USA, in 1987 and 1992, respectively. In August 1992, he joined the Department of Electrical Engineering, National Chung Hsing University, Taiwan, where he is currently a Professor. His research expertise lies in information-theoretical image processing with applications in medical imaging and remote sensing. He has significant experience in developing hyperspectral imaging models for magnetic resonance (MR) imaging applications and in the software engineering aspects of implementing and testing these methods. Furthermore, his research interests extend to the development and evaluation of cloud computing systems for various real-world applications. He also integrates artificial intelligence, machine learning, cognitive science, and deep neural networks into his work, applying these to intelligent systems such as decision support systems, human-computer interaction, and machine learning, as well as to algorithmic problems aimed at improving efficiency. His diverse research interests include: hyperspectral image processing, medical image processing, agricultural production applications, communication networks, network security in mobile networks, cloud computing, multimedia system design, and performance evaluation.
\end{IEEEbiography}

\begin{IEEEbiography}[{\includegraphics[width=1in,height=1.25in,clip,keepaspectratio]{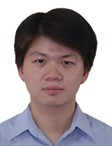}}]{Tsung-Jung Liu}
(Member, IEEE) received the B.S. degree in electrical engineering from National Tsing Hua University, Hsinchu, Taiwan, in 1998, the M.S. degree in communication engineering from National Taiwan University, Taipei, Taiwan, in 2001, and the Ph.D. degree in electrical engineering from the University of Southern California, Los Angeles, CA, USA, in 2014. He is currently an Associate Professor with the Department of Electrical Engineering, and the Graduate Institute of Communication Engineering, National Chung Hsing University, Taichung, Taiwan. His research interests include computer vision, perceptual image/video processing, visual quality assessment, image restoration, machine learning/deep learning, and artificial intelligence.
\end{IEEEbiography}

\end{document}